\documentclass[10pt,twocolumn,letterpaper]{article}

\usepackage[pagenumbers]{cvpr}      

\usepackage{booktabs}
\usepackage{graphicx}
\usepackage{amsmath,amssymb}
\usepackage{xcolor}

\definecolor{cvprblue}{rgb}{0.21,0.49,0.74}
\usepackage[pagebackref,breaklinks,colorlinks,allcolors=cvprblue]{hyperref}

\def\confName{CVPR}
\def\confYear{2027}

\title{Where the Image Sits: Vision--Text Order Bias in Jev}

\author{Youzhi Zhang\\
 Hong Kong Institute of Science \& Innovation, Chinese Academy of Sciences}

\begin{document}
\maketitle
\begin{abstract}
Jev-style system-1 decision models read a choice from a single prefill. They restrict the next-token distribution to the supplied option labels and return a distribution over the options, with no generation and no parsing. Recent work shows that generative vision-language models are sensitive to the relative order of image and text tokens. We ask whether that sensitivity reaches this Jev-style readout. The answer is always read at the end of the prompt, so moving the image changes the distance from the visual evidence to the decision, but never the position at which the decision is read. Using four placements---prefix, pre-question, pre-options, and suffix---and holding the option listing fixed, we measure argmax flips, total variation, accuracy, calibration, and label marginals.
Suffix placement flips 41.0\% of decisions and lowers accuracy by 18.6 points. Averaged over the four slots, the curve is monotone, not U-shaped. That shape differs from prior results on generative vision-language models. The effect survives a battery of controls. 
We also introduce prefix--question averaging (P+Q), a two-forward readout that averages the prefix and pre-question distributions. It is at least as accurate as prefix and outperforms the prior baselines.  Our code and data are publicly available at
\url{https://github.com/Youzhi333/Visual_Tex_Order_Bias_Jev}.
\end{abstract}

\section{Introduction}
\label{sec:intro}

A Jev-style model offers a system-1 way to make a choice \citep{typesafe2026jev,almeida2026jev,anyjev2026}. Nothing is decoded: the model assigns the decision to the offered options in a single pass.
Visual Jev brings the same interface to a shared visual context~\citep{yu2026visualjev}. PixelJev takes a different route: an image, an instruction, and a runtime candidate set become a structured choice, with probabilities conditioned on the candidates~\citep{zhou2026pixeljev}. Operationally, the decision is a single prefill, with neither generation nor a parser.


Recent work establishes a \emph{vision--text order bias} in generative
vision-language models~\citep{gan2026order}. Placing the visual tokens last
(suffix order) underperforms placing them first (prefix order), and sweeping the
image position produces a U-shaped accuracy curve: both ends beat the middle,
consistent with a lost-in-the-middle failure. The proposed remedy is a
dual-order contrastive decoder that runs a prefix-order teacher and an
order-$r$ student and fuses their token distributions. We use ``bias''
operationally, as a change in decision relative to a placement-invariant
average of the image positions. It is not a claim that prefix is the correct
order: the option listing is fixed and never rotated, and the invariant
reference averages only image placements.

A Jev-style system-1 readout is a different inference regime. It returns one distribution over the option labels, so there is no decoding step and no sampling variance. The answer position is always the end of the prompt. Moving the image therefore changes the distance from the visual evidence to the decision, but it never changes where the decision is read. This yields two competing predictions. Suffix order may be \emph{better} than in generation, because the image sits next to the answer. Or prefix order may remain better, because that is the order the model was trained on. Which prediction holds is the question this paper asks.

\begin{figure*}[t]
\centering
\includegraphics[width=\textwidth]{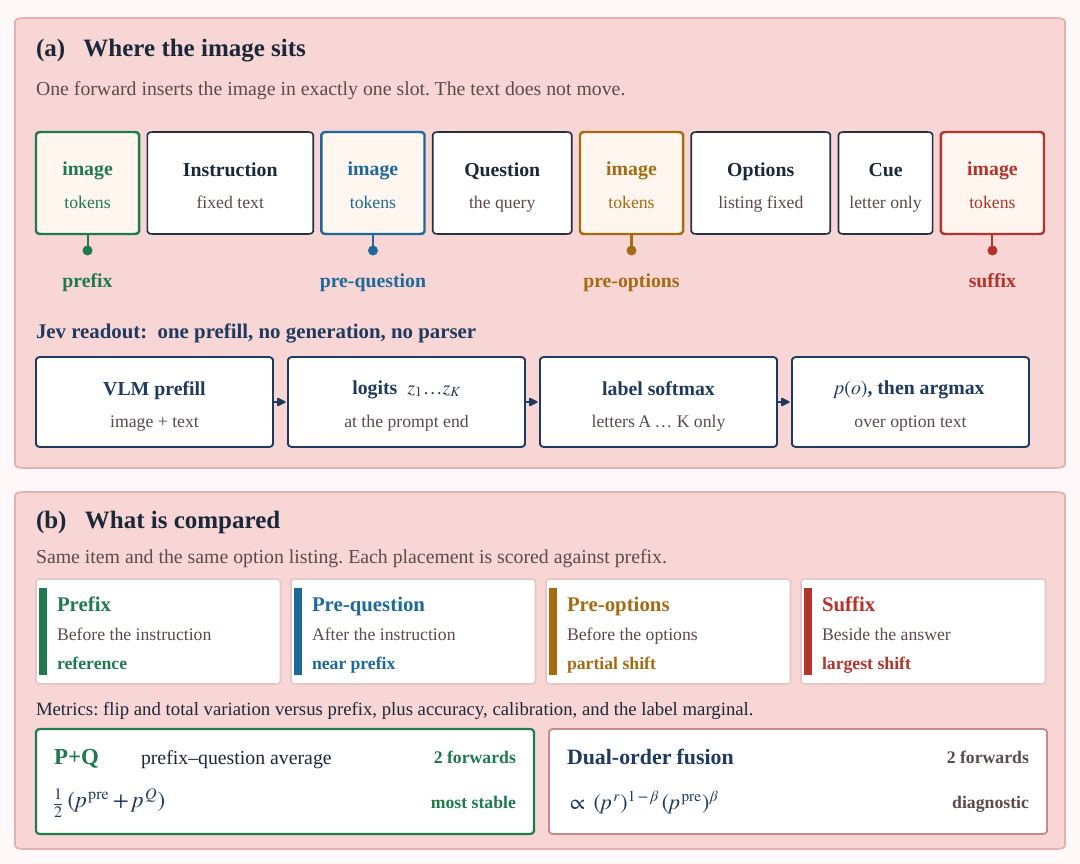}
\caption{Framework. (a) The image block is inserted at one of four slots---prefix, pre-question, pre-options, or suffix---while the instruction, question, option listing, and answer cue stay fixed. A Jev readout is one prefill: logits at the end of the prompt are softmaxed over the option labels only, and the decision is the argmax. (b) Each placement is compared with prefix by flip, total variation, accuracy, calibration, and the label marginal. P+Q averages the prefix and pre-question distributions. Dual-order fusion  is used as a diagnostic.}
\label{fig:framework}
\end{figure*}

We study the modality axis in isolation, as shown in Figure \ref{fig:framework}. The main sweep uses four placements---prefix, pre-question, pre-options, and suffix---with the option listing fixed per item. We measure the fraction of decisions that flip, the total variation between distributions, accuracy, calibration, and the label marginal. We then repeat the sweep across several fixed listings and under pointwise scoring \cite{huang2026runoff,zhang2026ijev}, which is listing-invariant by construction. A grey-image control separates a positional effect from a visual-content effect, and a label-permutation control separates option identity from letter identity. Finally, we define a training-free \emph{prefix--question averaging} readout (P+Q), which averages the prefix and pre-question distributions in two forwards per item, and we test a training-free \emph{dual-order readout} that fuses the prefix and order-$r$ option logits \citep{gan2026order}. P+Q is the strongest stabilizer in the sweep; dual-order fusion is a diagnostic.  

\paragraph{Contributions.}
\begin{itemize}
\item A characterization of vision--text position sensitivity in a Jev-style system-1 decision readout, as opposed to generative decoding. Suffix placement flips 41.0\% of decisions and lowers accuracy by 18.6 points. The four-slot curve is monotone on average, not U-shaped. That shape differs from prior results on generative vision-language models.
\item A controlled sweep of image position with the option order held fixed, reporting decision flips and total variation alongside accuracy, paired tests, and a placement-invariant reference. The effects above survive this battery of controls.
\item Robustness controls---repeated listings, pointwise scoring, a grey image, and label permutation. The robust claim is decision stability rather than a uniform accuracy loss: the sign of the accuracy change remains dataset-dependent.
\item Prefix--question averaging (P+Q), a two-forward placement average that is at least as accurate as prefix, with a 2.2--2.4\% flip, together with the other placement-subset ablations. A monotone decline does not mean that prefix alone is best. On the all-model mean, P+Q outperforms dual-order fusion.
\end{itemize}

\section{Related Work}
\label{sec:related}


\paragraph{Vision--text order.}
\citet{gan2026order} vary the relative position of visual and text tokens and
find that suffix order underperforms prefix order, with a U-shaped accuracy
curve across image positions. Their dual-order contrastive decoder is
generative: it runs a prefix teacher and an order-$r$ student, keeps the top-$k$
student candidates, and scores each candidate with a convex combination of the
two scores before greedy decoding. We adopt their position parameter and their
prefix/suffix/middle distinction, but apply them to a one-prefill label
readout. The readout has no decoding step, so the natural analogue is the same
convex score combination restricted to the $K$ option log-probabilities; with
log-probability scores it is Eq.~(\ref{eq:fuse}), and with probability scores it
is the prefix-suffix probability mean, which we also report as a baseline. We
evaluate both score rules but do not run DOCD's top-$k$/greedy loop, because
there is nothing to decode. Related work finds beginning/end
preferences in multimodal contexts~\citep{tan2024order}, documents both choice
and modality ordering across LVLMs~\citep{wei2026order}, audits ordering facets
with large per-facet flip rates~\citep{paruchuri2026same}, reports large
per-cell flips in video multiple-choice while scores move
little~\citep{zhu2026stable}, varies option IDs and delimiters across
formats~\citep{rosenthal2026unexplored}, and treats option permutation as a
simple attack~\citep{zong2024fool}. Taken together, they show that ordering can
have a substantial effect. None isolates the modality axis in a decision
readout, which is the question here.


\paragraph{Option order and position.}
In language models, reordering the choices can change the selected answer, and
the chance of such a change rises with the number of
choices~\citep{zheng2024large,pezeshkpour-hruschka-2024-large,zhang2026visualjevoptionbiaszhang2026}. Averaging
log-probabilities across cyclic rotations (L0) cancels an additive position
term exactly~\citep{zheng2024large,anyjev2026,zhang2026visualjevoptionbiaszhang2026}. Other routes to order
independence include calibration~\citep{mcilroyyoung2024orderindependence}, independent option scoring, and a position-balanced method~\citep{huang2026runoff,zhang2026ijev}. Those effects are real, but they are not the subject of this paper. We hold the option listing fixed and study a different axis: vision--text order.
We use the pointwise readout as a listing-invariant control, not as a proposed
method, and we do not average over option rotations in the main sweep.

\paragraph{Calibration and priors.}
Calibration can remove a content-free prior estimated from
probes~\citep{zhao2021calibrate}; a batch prior comes from the current batch
instead~\citep{zhou2024batch}. Both assume the model's label marginal is a bias
rather than signal. We report calibration and label marginals as diagnostics;
we do not divide a prior in the main readout, because on per-item VQA each
question is asked once and the per-question prior is not estimable. Calibration
of neural networks~\citep{guo2017calibration} and paired bootstrap confidence
intervals~\citep{efron1979bootstrap} are the statistical tools we use.

\section{Modality Order in Jev}
\label{sec:method}

\paragraph{Setup.}
A state, written $x$, consists of a set of images $I$ and, when present, text $T$. For a question $q$, the available choices form $\mathcal{O}=\{o_1,\dots,o_K\}$, and their presentation order is the caller's listing $\tau=(\tau_1,\dots,\tau_K)$. The rendered prompt contains the images, an instruction, the question, and the labeled options $\ell_1,\dots,\ell_K\in\{\mathrm{A},\dots,\mathrm{Z}\}$. The model must return only the label. A single prefill yields the logits $z_1,\dots,z_K$ at the position where the answer is read. The readout is
\begin{equation}
  p_\tau(o_{\tau_j}) = \frac{e^{z_j}}{\sum_{k=1}^{K} e^{z_k}},
  \label{eq:readout}
\end{equation}
a distribution over the option texts. Nothing is generated and nothing is parsed. The subscript $\tau$ denotes the listing. Except where noted, comparisons align probability mass by option text rather than by slot.

\paragraph{Modality placement.}
Let the text-token sequence be $e=[x_1,\dots,x_L]$ and let $v$ be the block of image tokens. For a relative position $r\in[0,1]$,
\begin{equation}
  s(r) = [x_1,\dots,x_{\lfloor rL\rfloor},\, v,\, x_{\lfloor rL\rfloor+1},\dots,x_L].
  \label{eq:position}
\end{equation}
Following \citet{gan2026order}, prefix is $r=0$ (image first) and suffix is $r=1$ (image last); any intermediate $r$ is a middle placement. We use four semantic slots, in increasing token position: prefix (the image before the instruction), pre-question (between the instruction and the question), pre-options (between the question and the option list), and suffix (after the options). They are implemented with the chat template's content parts. A token-splice path is available for exact intermediate positions. Appendix~\ref{sec:app-prompts} gives the templates and the validation checks.

\paragraph{Option-order control.}
The option listing is fixed per item in the main sweep. We repeat the sweep over three to five fixed listings and under a pointwise readout, which is listing-invariant by construction \citep{huang2026runoff,zhang2026ijev}. 

\paragraph{Dual-order fusion.}
Let $p^{pre}$ be the prefix readout ($r=0$) and $p^{r}$ the readout at position
$r$. For $\beta\in[0,1]$ define
\begin{equation}
  p^{fuse}_\beta(o) \;\propto\;
  \bigl(p^{r}(o)\bigr)^{1-\beta}\bigl(p^{pre}(o)\bigr)^{\beta}.
  \label{eq:fuse}
\end{equation}
$\beta=0$ returns the student, $\beta=1$ the teacher, and $\beta=0.5$ is the log-mean. This is a readout-time analogue of the DOCD score rule~\citep{gan2026order}. DOCD is a generative decoder with a top-$k$/greedy loop, but its per-step convex combination, restricted to the $K$ option labels, is the same rule as Eq.~\ref{eq:fuse}. The readout costs two forwards per item and samples nothing; we do not run the top-$k$/greedy loop. We sweep $\beta\in\{0,0.25,0.5,0.75,1\}$ offline from cached prefix and order-$r$ logits, so the sweep itself adds no forwards.

\paragraph{Prefix--question averaging (P+Q).}
Write $p^{pre}$, $p^{Q}$, $p^{O}$ and $p^{S}$ for the readouts at prefix,
pre-question, pre-options and suffix. For a subset $A$ of these placements,
define the placement average
\begin{equation}
  p^{A}(o) \;=\; \frac{1}{|A|}\sum_{r\in A} p^{r}(o).
  \label{eq:placement-avg}
\end{equation}
The P+Q readout is the case $A=\{pre,Q\}$:
\begin{equation}
  p^{P+Q}(o) \;=\; \tfrac12\bigl(p^{pre}(o)+p^{Q}(o)\bigr),
  \hat o \;=\; \arg\max_{o\in\mathcal{O}} p^{P+Q}(o).
  \label{eq:pq}
\end{equation}
It is a linear opinion pool over the two prefix-side placements: a valid
distribution over the same option texts, computed under the same listing and
label restriction as the single-placement readout. It costs two forwards per
item and samples nothing. It is not placement-invariant---it averages two
specific prefix-side placements and never reads the suffix---so it is a
placement policy, not a correction for a prompt whose image is fixed at suffix;
like every multi-placement readout, it requires the caller to render both
placements. The other subsets in Eq.~(\ref{eq:placement-avg}) are reported as
ablations in Section~\ref{sec:results}, which compares P+Q with prefix,
dual-order fusion and the all-position mean.

\paragraph{Metrics.}
Let $p_{r,n}$ be the distribution for item $n$ at position $r$, aligned by
option, and let $p_{0,n}$ be prefix. The \emph{modality flip} is
\begin{equation}
  F(r) = \frac{1}{N}\sum_{n=1}^{N}
  \mathbf{1}\!\left[\arg\max_o p_{r,n}(o) \neq \arg\max_o p_{0,n}(o)\right],
  \label{eq:flip}
\end{equation}
and the \emph{modality TV} is
\begin{equation}
  T(r) = \frac{1}{N}\sum_{n=1}^{N}\frac12\sum_o
  \left|p_{r,n}(o)-p_{0,n}(o)\right|.
  \label{eq:tv}
\end{equation}
\emph{Instability} is the mean flip over all unordered pairs of positions.
\emph{Accuracy} is the fraction of items whose argmax is the gold option. We
report expected calibration error (ECE) over ten confidence bins
\citep{guo2017calibration} and the label marginal, i.e., the mean probability of
each letter across items~\citep{zhao2021calibrate,zhou2024batch}. The
\emph{cost} is the number of forwards per item: one for a single readout, two
for dual-order fusion, and $K$ for pointwise scoring.

\paragraph{Controls.}
We run a battery of controls, each aimed at one alternative explanation.
(i) \emph{Listing robustness}: repeat the sweep over several fixed listings and
report the spread of $F(r)$ and $T(r)$. (ii) \emph{Pointwise}: score each option
alone with a yes/no question and read the yes/no log-odds; the result is
listing-invariant, so a surviving effect cannot be an option-order artifact.
(iii) \emph{Grey image}: replace the image with a uniform grey field of the same
size, the image analogue of a content-free probe~\citep{zhao2021calibrate}; a
residual effect is positional or textual, not visual-content-driven. (iv)
\emph{Label permutation}: permute the letter-to-option assignment while holding
the option order fixed, following label-prior
controls~\citep{zhao2021calibrate,zhou2024batch}; a change in flips isolates
letter preference from option identity. (v) \emph{Placement-invariant
reference}: compare each placement with the per-item mean of the four placement
distributions, which gives no slot priority; this is an image-position average,
not an option-order average, and no option rotation is used. (vi) \emph{Token
distance}: measure the number of text tokens from the last image token to the
answer position, and add a fixed-distance pair that moves the image across the
option list while matching that distance exactly. (vii) \emph{Full-set
reruns}: repeat the comparison on the entire A-OKVQA, EuroSAT and Pets splits,
not on the 300-item subsets. (viii) \emph{Readout baselines}: compare against
training-free averages and fusions of the cached placement distributions,
including all-position mean and log-mean, prefix--suffix mean, placement
subsets, P+Q and dual-order fusion. (ix) \emph{Paired tests}: test the per-item
suffix-minus-prefix accuracy difference with cell-bootstrap intervals and exact
McNemar tests on discordant items. (x) \emph{Downstream proxy}: gate actions on
confidence, report utility, coverage and the area under the risk--coverage curve
(AURC), and repeat after temperature-scaling each position.
These controls target, respectively, listing order, option comparison, visual
content, letter identity, the choice of reference, token distance, subset sampling,
readout choice, item pairing, and downstream utility.

\begin{table*}[t]
\centering
\footnotesize

\begin{tabular}{@{}lrrrrrr rrrr ccc@{}}
\toprule
& \multicolumn{3}{c}{Flip vs prefix} & \multicolumn{3}{c}{TV vs prefix} & \multicolumn{4}{c}{Accuracy} & \multicolumn{3}{c}{Readout accuracy} \\
\cmidrule(lr){2-4}\cmidrule(lr){5-7}\cmidrule(lr){8-11}\cmidrule(l){12-14}
Dataset & $F_{\mathrm{mid}}$ & $F_{\mathrm{ques}}$ & $F_{\mathrm{suf}}$ & $T_{\mathrm{mid}}$ & $T_{\mathrm{ques}}$ & $T_{\mathrm{suf}}$ & prefix & pre-Q & pre-O & suffix & P+Q & DOF 0.5 & DOF 0.75 \\
\midrule
\multicolumn{14}{l}{\textit{All-model mean}}\\
A-OKVQA & 0.019 & 0.110 & 0.350 & 0.019 & 0.116 & 0.365 & 0.868 & 0.865 & 0.826 & 0.615 & 0.868 & 0.810 & 0.859 \\
ScienceQA & 0.053 & 0.127 & 0.264 & 0.044 & 0.122 & 0.251 & 0.870 & 0.867 & 0.819 & 0.719 & 0.874 & 0.865 & 0.869 \\
CV-Bench & 0.026 & 0.281 & 0.387 & 0.024 & 0.227 & 0.323 & 0.798 & 0.800 & 0.680 & 0.598 & 0.799 & 0.765 & 0.793 \\
EuroSAT & 0.070 & 0.249 & 0.456 & 0.071 & 0.224 & 0.380 & 0.462 & 0.470 & 0.484 & 0.381 & 0.471 & 0.440 & 0.463 \\
Pets & 0.027 & 0.089 & 0.623 & 0.028 & 0.088 & 0.588 & 0.728 & 0.724 & 0.730 & 0.369 & 0.725 & 0.689 & 0.731 \\
MMMU & 0.054 & 0.252 & 0.380 & 0.038 & 0.191 & 0.305 & 0.520 & 0.527 & 0.503 & 0.448 & 0.526 & 0.513 & 0.524 \\
Mean & 0.042 & 0.185 & 0.410 & 0.037 & 0.161 & 0.369 & 0.708 & 0.709 & 0.674 & 0.522 & 0.711 & 0.680 & 0.706 \\
\midrule
\multicolumn{14}{l}{\textit{Qwen mean}}\\
A-OKVQA & 0.022 & 0.079 & 0.167 & 0.020 & 0.080 & 0.175 & 0.865 & 0.862 & 0.850 & 0.793 & 0.865 & 0.852 & 0.863 \\
ScienceQA & 0.057 & 0.092 & 0.188 & 0.048 & 0.085 & 0.176 & 0.888 & 0.882 & 0.862 & 0.792 & 0.892 & 0.880 & 0.888 \\
CV-Bench & 0.020 & 0.193 & 0.301 & 0.023 & 0.163 & 0.283 & 0.846 & 0.846 & 0.772 & 0.679 & 0.847 & 0.833 & 0.847 \\
EuroSAT & 0.093 & 0.256 & 0.344 & 0.084 & 0.234 & 0.312 & 0.479 & 0.492 & 0.508 & 0.485 & 0.492 & 0.509 & 0.507 \\
Pets & 0.018 & 0.073 & 0.373 & 0.021 & 0.076 & 0.383 & 0.843 & 0.846 & 0.843 & 0.612 & 0.843 & 0.824 & 0.853 \\
MMMU & 0.048 & 0.161 & 0.279 & 0.037 & 0.141 & 0.239 & 0.552 & 0.556 & 0.568 & 0.528 & 0.553 & 0.557 & 0.553 \\
Mean & 0.043 & 0.142 & 0.276 & 0.039 & 0.130 & 0.261 & 0.745 & 0.747 & 0.734 & 0.648 & 0.749 & 0.743 & 0.752 \\
\midrule
\multicolumn{14}{l}{\textit{Non-Qwen mean}}\\
A-OKVQA & 0.013 & 0.152 & 0.593 & 0.017 & 0.163 & 0.617 & 0.871 & 0.868 & 0.793 & 0.379 & 0.872 & 0.753 & 0.852 \\
ScienceQA & 0.048 & 0.173 & 0.364 & 0.039 & 0.171 & 0.352 & 0.847 & 0.847 & 0.762 & 0.621 & 0.849 & 0.844 & 0.842 \\
CV-Bench & 0.034 & 0.399 & 0.502 & 0.026 & 0.314 & 0.377 & 0.734 & 0.738 & 0.558 & 0.490 & 0.736 & 0.673 & 0.722 \\
EuroSAT & 0.040 & 0.240 & 0.604 & 0.053 & 0.211 & 0.470 & 0.440 & 0.441 & 0.453 & 0.243 & 0.444 & 0.348 & 0.404 \\
Pets & 0.039 & 0.110 & 0.957 & 0.037 & 0.104 & 0.861 & 0.574 & 0.562 & 0.578 & 0.043 & 0.567 & 0.508 & 0.569 \\
MMMU & 0.061 & 0.373 & 0.513 & 0.040 & 0.258 & 0.392 & 0.479 & 0.488 & 0.418 & 0.341 & 0.490 & 0.453 & 0.484 \\
Mean & 0.039 & 0.241 & 0.589 & 0.035 & 0.203 & 0.511 & 0.658 & 0.657 & 0.594 & 0.353 & 0.660 & 0.597 & 0.646 \\
\bottomrule
\end{tabular}
\caption{Modality flip $F$ and total variation $T$ relative to prefix, accuracy
at the four image placements, and accuracies of the P+Q mean, dual-order fusion
at $\beta=0.5$ and $0.75$, for the all-model mean over seven
checkpoints (top), the Qwen mean (middle) and the non-Qwen mean (bottom).
Subscripts mid, ques and suf denote pre-question, pre-options and suffix;
``Readout accuracy'' columns use the same rows.   Per-cell values for all seven checkpoints, including
pointwise, are in Table~\ref{tab:positionsfull}.}
\label{tab:positions}
\end{table*}

\section{Experimental Setup}
\label{sec:setup}
\paragraph{Models.}
The primary models are Qwen3-VL-2B/4B/8B-Instruct \citep{qwen3vl2025} and Qwen2.5-VL-7B-Instruct \citep{wang2024qwen2vl}, in bfloat16, with the image-token budget capped at $448^2$ pixels. Batches contain four items for the 2B checkpoint and one item for every larger checkpoint. Runs are single-GPU on an RTX 4090 D. The full suite is repeated on three non-Qwen checkpoints: Idefics3-8B \citep{laurencon2024idefics3}, LLaVA-OneVision-7B \citep{li2024llavaonevision}, and LLaVA-NeXT-7B \citep{liu2024llavanext}. 

\paragraph{Datasets.}
Table~\ref{tab:data} in the appendix lists the evaluation sets. A-OKVQA \citep{schwenk2022aokvqa}, ScienceQA \citep{lu2022scienceqa}, and CV-Bench \citep{tong2024cambrian} are per-item VQA with short option lists. EuroSAT \citep{helber2019eurosat} and Pets~\citep{parkhi2012pets} use one fixed option list for every item, so class and slot are aliased. MMMU \citep{yue2024mmmu} has 2--9 options and often several images. The 26-option cap is the A--Z limit of the letter readout.

\paragraph{Protocol.}
The main grid uses 300-item subsets of all six datasets, four semantic placements (prefix, pre-question, pre-options, and suffix), the four Qwen models, and the three non-Qwen checkpoints on the same grid. Every item is decided once per placement. The readout is deterministic, and no tokens are sampled. Accuracy claims are confirmed on the full A-OKVQA, EuroSAT, and Pets sets at prefix, pre-question, and suffix. The option listing is fixed per item in the main sweep. Listing robustness, pointwise scoring, the grey-image control, and label permutation use 300-item subsets; label permutation covers all seven checkpoints on all six grids. Every run records the library versions, batch size, dtype, and driver in the result JSON. 
Appendix \ref{sec:app-prompts} shows the implementation details.

\begin{figure*}[t]
\centering
\includegraphics[width=\textwidth]{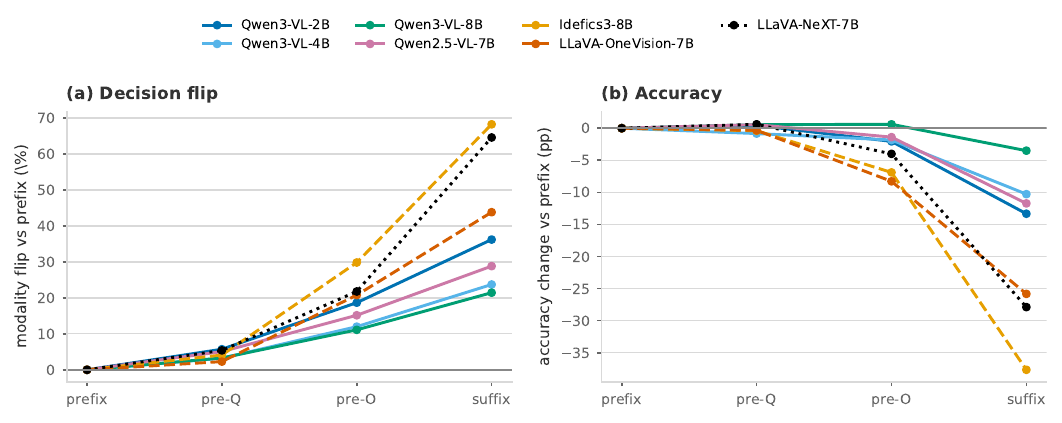}
\caption{The four-slot position sweep for every checkpoint, averaged over its
six datasets: (a) modality flip relative to prefix; (b) accuracy change
relative to prefix in percentage points. Each line is one checkpoint and is
plotted relative to its own prefix, so all lines start at zero; there are no
error bars. Suffix flips span 21.4--68.3\% and accuracy losses 3.5--37.6 points;
dashed and dotted lines are the non-Qwen checkpoints. The four-slot curve is
monotone for all seven models.}
\label{fig:family}
\end{figure*}

\begin{figure}[t]
\centering
\includegraphics[width=\columnwidth]{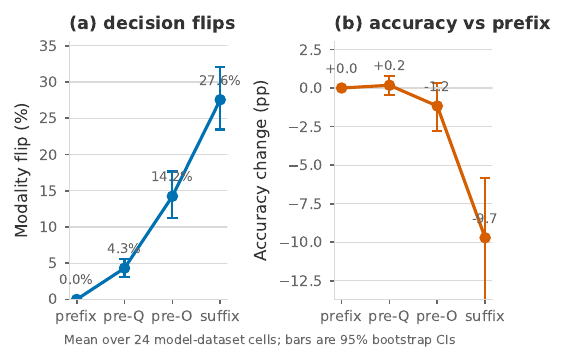}
\caption{Position sweep, mean over 24 Qwen model--dataset cells (four models
$\times$ six datasets). (a) Modality flip relative to prefix; (b) accuracy
change relative to prefix in percentage points. Pre-Q: the image between the
instruction and the question; pre-O: between the question and the options. Bars are 95\%
bootstrap intervals over cells, not items. Prefix is zero by construction in
(a) and by definition in (b); the suffix values are 27.6\% and $-9.7$ points.
The mean curve is monotone, but individual cells need not be.}
\label{fig:curve}
\end{figure}

\section{Results}
\label{sec:results}

Table~\ref{tab:positions} reports dataset means for all seven checkpoints (top), the four Qwen checkpoints (middle), and the three non-Qwen checkpoints (bottom). $F_{\mathrm{mid}}$, $F_{\mathrm{ques}}$, and $F_{\mathrm{suf}}$ are the fractions of decisions that differ from prefix at pre-question, pre-options, and suffix; despite the subscript, $F_{\mathrm{mid}}$ is pre-question, not a midpoint. $T_{\mathrm{mid}}$, $T_{\mathrm{ques}}$, and $T_{\mathrm{suf}}$ are the matching total-variation distances. The next four columns are accuracies at prefix, pre-question, pre-options, and suffix, and the last three are P+Q and dual-order fusion at $\beta=0.5$ and $0.75$. Mean rows weight each model--dataset cell equally, not each item.

Table~\ref{tab:fuse} compares readouts on the same three panels. Each row is one readout, and each panel reports accuracy, flip relative to prefix, ECE, and forwards per item. Apart from pointwise, every row is computed offline from the four cached placement distributions, so the differences are properties of the readout, not new model runs. In the averaging rows, P, Q, and O abbreviate prefix, pre-question, and pre-options. All-mean is the mean of the four placement distributions, All-logmean their log-mean, P+S the prefix--suffix probability mean, and DOF~$\beta$ dual-order fusion at $\beta$. Pointwise scores each option alone with a yes/no question, so the decision cannot be a listing artifact. It costs $K$ forwards and is run only on A-OKVQA and EuroSAT. In Table~\ref{tab:fuse} its accuracy falls from 0.662 to 0.524, with a 42.3\% flip (29.7\% on Qwen, 58.9\% on non-Qwen), so the effect survives listing-invariant scoring.

Per-cell values are in Tables~\ref{tab:positionsfull} and~\ref{tab:fusefull} of the appendix. Averaging ablations are in Table~\ref{tab:avgfull}, and checkpoint and dataset aggregates are in Tables~\ref{tab:bymodel} and~\ref{tab:bydataset} of the appendix. Appendix \ref{app:sec_baselines} discusses baselines in detail, and Appendix \ref{app:sec_controls} shows more results on controls.

\begin{table*}[t]
\centering
\footnotesize
\begin{tabular}{@{}l rrrr rrrr rrrr@{}}
\toprule
& \multicolumn{4}{c}{\textit{All-model mean}} & \multicolumn{4}{c}{\textit{Qwen mean}} & \multicolumn{4}{c}{\textit{Non-Qwen mean}} \\
\cmidrule(lr){2-5}\cmidrule(lr){6-9}\cmidrule(l){10-13}
Readout & Acc. & Flip & ECE & Fwd & Acc. & Flip & ECE & Fwd & Acc. & Flip & ECE & Fwd \\
\midrule
Prefix & 0.708 & 0.000 & 0.146 & 1 & 0.745 & 0.000 & 0.159 & 1 & 0.658 & 0.000 & 0.127 & 1 \\
Pre-question & 0.709 & 0.042 & 0.150 & 1 & 0.747 & 0.043 & 0.163 & 1 & 0.657 & 0.039 & 0.132 & 1 \\
Pre-options & 0.674 & 0.185 & 0.167 & 1 & 0.734 & 0.142 & 0.169 & 1 & 0.594 & 0.241 & 0.163 & 1 \\
Suffix & 0.522 & 0.410 & 0.237 & 1 & 0.648 & 0.276 & 0.220 & 1 & 0.353 & 0.589 & 0.260 & 1 \\
All-position mean & 0.709 & 0.073 & 0.113 & 4 & 0.751 & 0.067 & 0.118 & 4 & 0.652 & 0.082 & 0.105 & 4 \\
All-position log-mean & 0.705 & 0.089 & 0.130 & 4 & 0.751 & 0.080 & 0.150 & 4 & 0.643 & 0.102 & 0.103 & 4 \\
Prefix+suffix mean & 0.685 & 0.130 & 0.134 & 2 & 0.743 & 0.106 & 0.130 & 2 & 0.606 & 0.161 & 0.139 & 2 \\
P+Q+O mean & 0.711 & 0.049 & 0.129 & 3 & 0.752 & 0.043 & 0.138 & 3 & 0.656 & 0.058 & 0.117 & 3 \\
Q+O mean & 0.707 & 0.091 & 0.131 & 2 & 0.750 & 0.080 & 0.136 & 2 & 0.649 & 0.105 & 0.124 & 2 \\
P+Q mean & 0.711 & 0.023 & 0.145 & 2 & 0.749 & 0.024 & 0.157 & 2 & 0.660 & 0.022 & 0.129 & 2 \\
P+O mean & 0.708 & 0.082 & 0.129 & 2 & 0.753 & 0.073 & 0.134 & 2 & 0.649 & 0.095 & 0.122 & 2 \\
M+M mean & 0.708 & 0.082 & 0.115 & 4 & 0.752 & 0.075 & 0.129 & 4 & 0.648 & 0.091 & 0.096 & 4 \\
Dual-order, $\beta{=}0.5$ & 0.680 & 0.145 & 0.136 & 2 & 0.743 & 0.112 & 0.152 & 2 & 0.597 & 0.188 & 0.116 & 2 \\
Dual-order, $\beta{=}0.75$ & 0.706 & 0.058 & 0.128 & 2 & 0.752 & 0.051 & 0.148 & 2 & 0.646 & 0.066 & 0.102 & 2 \\
Pointwise (prefix) & 0.662 & 0.000 & 0.178 & $K$ & 0.667 & 0.000 & 0.159 & $K$ & 0.657 & 0.000 & 0.203 & $K$ \\
Pointwise (suffix) & 0.524 & 0.423 & 0.143 & $K$ & 0.642 & 0.297 & 0.158 & $K$ & 0.366 & 0.589 & 0.122 & $K$ \\
\bottomrule
\end{tabular}
\caption{Readout variants and controls. Each row is one readout; the three
column groups are the all-model mean (42 model--dataset cells), the Qwen mean
(24) and the non-Qwen mean (18), each with accuracy, flip relative to prefix,
ECE and forwards per item. 
All-mean is the
mean of the four placement
distributions, All-logmean their log-mean, P+S mean the prefix--suffix
probability mean, and DOF $\beta$ dual-order fusion at $\beta$. In the averaging
rows P is prefix, Q is pre-question, O is pre-options, and M+M is the mean of
the all-position mean and all-position log-mean distributions; each averages
the distributions of the named placements. The all-position mean is the
placement-invariant reference; $\beta$ is not tuned per cell. Prefix is the
single-placement default. Per-cell values are in Table~\ref{tab:fusefull}.}
\label{tab:fuse}
\end{table*}

\paragraph{Image position changes the decision.}
On the all-model panel of Table~\ref{tab:positions}, the flip is 4.2\% at pre-question, 18.5\% at pre-options, and 41.0\% at suffix; total variation is 0.037, 0.161, and 0.369; and accuracy is 0.708, 0.709, 0.674, and 0.522 (Figures~\ref{fig:family}--\ref{fig:curve}). Moving the image from prefix to suffix therefore changes 41\% of decisions and costs 18.6 points. Pre-question is essentially free (0.709), while pre-options already costs 3.4 points. The increase is monotone on average, not U-shaped. The families agree in direction and differ in magnitude: Qwen flips 27.6\% and loses 9.7 points, and non-Qwen flips 58.9\% and loses 30.5 points. P+Q and dual-order fusion both recover prefix-level accuracy. On the all-model mean, P+Q is more accurate than both the prefix readout and dual-order fusion.

\paragraph{The average curve is monotone.}
Figure~\ref{fig:curve} aggregates the 24 Qwen model--dataset cells. Panel (a) is the flip relative to prefix, and panel (b) is the accuracy change in percentage points, with 95\% bootstrap intervals over cells. The suffix point is a 27.6\% flip and a $-9.7$ point change, so on the average the flip rises and accuracy falls monotonically. Panel (a) is zero at prefix by construction, and panel (b) is zero there by definition. Figure~\ref{fig:family} repeats the sweep for each checkpoint, averaged over its six datasets and plotted relative to that checkpoint's own prefix. Qwen suffix flips span 21.4--36.2\% and non-Qwen flips span 43.8--68.3\%; accuracy losses span 3.5--37.6 points. The lines separate the families, but each line is an average over datasets.

\paragraph{Readouts trade stability for cost.}
Table~\ref{tab:fuse} asks whether a readout can recover prefix accuracy while depending less on position. The single-placement rows are the four distributions of Table~\ref{tab:positions}; the remaining rows combine them. Accuracy is the argmax accuracy of the readout, flip is the fraction of items whose prediction differs from the prefix prediction (zero for prefix by construction), ECE is the calibration error, and the last column is the number of forwards per item. In the all-model columns, suffix is 0.522 at a 41.0\% flip; the all-position mean is 0.709 at 7.3\%; P+Q is 0.711 at 2.3\%; P+Q+O is 0.711 at 4.9\%; P+O is 0.708 at 8.2\%; dual-order fusion at $\beta=0.75$ is 0.706 at 5.8\%; and prefix is 0.708 at 0. Averaging therefore recovers most of the prefix accuracy while cutting the flip by 5--18$\times$. P+Q is the most stable readout. P+O is the most accurate readout in the Qwen columns (0.753), but not overall, and dual-order fusion is intermediate at two forwards. Calibration is a separate axis: all-model ECE is 0.113 for the all-position mean, 0.128 for dual-order fusion at $\beta=0.75$, 0.129 for P+O, and 0.145 for P+Q, so the most stable readout is not the best calibrated. Prefix remains the cheapest strong default. Appendix~\ref{app:sec_baselines} discusses these baselines in more detail.

\begin{table}[t]
\centering
\small
\setlength{\tabcolsep}{2pt}
\resizebox{\columnwidth}{!}{
\begin{tabular}{lrrrr}
\toprule
Family & four-slot & leave-one-out & five-position & inva. acc \\
\midrule
Qwen & 0.220 [0.18,0.27] & 0.256 [0.22,0.30] & 0.188 [0.15,0.24] & 0.751 \\
Non-Qw & 0.544 [0.45,0.65] & 0.592 [0.49,0.69] & 0.544 [0.45,0.65] & 0.652 \\
All & 0.359 [0.29,0.43] & 0.400 [0.33,0.47] & 0.341 [0.27,0.42] & 0.709 \\
\bottomrule
\end{tabular}
}
\caption{Robustness of the image-position reference: four-slot average,
leave-one-out average of the other three placements, and five token-position
average. Values are suffix flips against the reference, with cell-bootstrap
intervals.  }
\label{tab:invariantrobust}
\end{table}

\begin{table}[t]
\centering
\footnotesize

\begin{tabular}{lrrrr}
\toprule
Model & prefix & pre-Q & pre-O & suffix \\
\midrule
Qwen3-VL-4B & 46 & 45 & 33 & 6 \\
Idefics3-8B & 45 & 44 & 32 & 5 \\
LLaVA-OneVision-7B & 45 & 43 & 31 & 5 \\
LLaVA-NeXT-7B & 51 & 46 & 33 & 4 \\
\bottomrule
\end{tabular}
\caption{Mean token distance from the last image token to the answer position,
50 A-OKVQA items. The four placements are not equally spaced: suffix is nearly
adjacent, pre-options is about 12 tokens farther, and prefix is about 45--51
tokens away. This distance is a confound and a candidate explanation, not a
separate manipulation.}
\label{tab:distance}
\end{table}

\paragraph{Placement-invariant reference.}
To check that the choice of prefix is not doing the work, we compare every placement with the mean of the four per-item distributions, the all-position mean in Table~\ref{tab:fuse}. Table~\ref{tab:invariantrobust} repeats the comparison under a leave-one-out reference and under a five-position token reference.
We call the reference placement-invariant because it gives no slot priority.
Based on the data from Tables~\ref{tab:fuse} and~\ref{tab:fusefull}, we could obtain that suffix flips 35.9\% of decisions against this reference (41.0\% against prefix). The family means are 22.0\% for Qwen (27.6\% against prefix), 62.5\% for Idefics3-8B (68.3\%), 50.3\% for the two LLaVA checkpoints (54.2\%), and 54.4\% for non-Qwen overall. The invariant readout is more accurate than suffix in every panel: 0.709 versus 0.522 all-model, 0.751 versus 0.648 on Qwen, and 0.652 versus 0.353 on non-Qwen. The non-Qwen row splits into 0.631 versus 0.268 for Idefics3 and 0.662 versus 0.396 for the two LLaVA checkpoints. The direction therefore survives the change of reference. Prefix-relative numbers overstate the magnitude by 4--6 points. Recomputing the reference over the five token-level placements gives 18.8\% for Qwen and 54.4\% for non-Qwen, consistent with the four-slot version. A leave-one-out reference, which averages only the other three placements, gives a larger suffix flip: 25.6\% for Qwen $[21.7,30.4]$ and 59.2\% for non-Qwen $[49.3,69.3]$. Including the compared placement in the average was therefore conservative. Per position, the leave-one-out flips are 10.7/9.8/12.2/25.6\% (Qwen) and 13.5/13.4/25.5/59.2\% (non-Qwen) at prefix, pre-question, pre-options, and suffix. Suffix is the outlier. Table~\ref{tab:invariantrobust} reports all three references with cell-bootstrap intervals. Each average is over image placements only, and none rotates the options.

\begin{table}[t]
\centering
\footnotesize
\setlength{\tabcolsep}{2pt}
\resizebox{\columnwidth}{!}{
\begin{tabular}{lrrrrr}
\toprule
Model & cells & label pre/suf & position id & position perm &  $p<.05$ \\
\midrule
Qwen3-VL-2B & 6 & 0.179/0.342 & 0.362 & 0.381 & 1/6 \\
Qwen3-VL-4B & 6 & 0.091/0.189 & 0.237 & 0.279 & 2/6 \\
Qwen3-VL-8B & 6 & 0.097/0.136 & 0.214 & 0.221 & 1/6 \\
Qwen2.5-VL-7B & 6 & 0.099/0.183 & 0.288 & 0.285 & 1/6 \\
Idefics3-8B & 6 & 0.244/0.824 & 0.683 & 0.669 & 1/6 \\
LLaVA-OneVision-7B & 6 & 0.150/0.448 & 0.438 & 0.460 & 2/6 \\
LLaVA-NeXT-7B & 6 & 0.226/0.627 & 0.646 & 0.587 & 1/6 \\
\midrule
Qwen & 24 & 0.117/0.212 & 0.276 & 0.291 & 5/24 \\
\midrule
Non-Qwen & 18 & 0.207/0.633 & 0.589 & 0.572 & 4/18 \\
\midrule
All & 42 & 0.155/0.393 & 0.410 & 0.412 & 9/42 \\
\bottomrule
\end{tabular}
}
\caption{Label-permutation 2$\times$2 for every checkpoint: option order fixed.
``label pre/suf'' are flips between identity and permuted letters at prefix and
suffix; ``position id/perm'' are suffix-vs-prefix flips under identity or
permuted letters. The position effect persists under permutation in every
family.  The last column is the interaction counts, which  are exact McNemar tests at $p<.05$.}
\label{tab:labelperm}
\end{table}

\paragraph{Token distance.}
Table~\ref{tab:distance} reports the number of text tokens between the last image token and the position where the answer is read. The four placements are not equally spaced: about 45--51 tokens at prefix, 43--46 at pre-question, 31--33 at pre-options, and 4--6 at suffix. The monotone flip and accuracy curves therefore track distance to the visual evidence. Because the study does not vary distance independently of the semantic slot, the results support a distance reading as much as a categorical-order reading. Separating the two would require a manipulation that holds token distance fixed while moving the image across a structural boundary.

\paragraph{Label permutation.}
Table~\ref{tab:labelperm} gives the summary $2\times 2$, and Table~\ref{tab:labelcells} the per-cell values. A label permutation reassigns the letters while holding the option order fixed. The label effect is the fraction of decisions that change between the identity labeling and the permuted labeling at a given placement. The position effect is the suffix-versus-prefix flip under either labeling, and the interaction tests whether that position effect itself changes under permutation. Across 42 cells, permutation flips 15.5\% of decisions at prefix and 39.3\% at suffix. The suffix label effect ranges from 13.6\% (Qwen3-VL-8B) to 82.4\% (Idefics3-8B), so letter identity is a real component of the readout. It does not explain the position effect: suffix-versus-prefix flips are 27.6\% under identity letters and 29.1\% under permuted letters on Qwen, and 58.9\% versus 57.2\% on non-Qwen. The paired interaction is significant in 9 of 42 cells, but the family-mean interaction is small and its interval includes zero: $+1.6$ points $[-0.3,3.1]$ for Qwen, $-1.7$ $[-4.0,0.7]$ for non-Qwen, and $+0.2$ $[-1.2,1.6]$ overall. Label identity and image position are separate effects. On the full A-OKVQA set the position flip persists across Qwen3-VL sizes, from 26.1\% to 27.9\% (2B), 15.4\% to 17.2\% (4B), and 12.3\% to 13.4\% (8B), with suffix label effects of 18.3\%, 10.7\%, and 7.7\%. For Qwen3-VL-4B, the EuroSAT and Pets full sets give 26.5\% to 26.5\% and 31.6\% to 30.1\%, with suffix label effects of 18.2\% and 19.8\%.

Appendix~\ref{app:sec_controls} reports the remaining controls: a confidence-gated downstream proxy, paired item-level tests, the grey-image control, full-set reruns, and token-level sweeps.

\paragraph{Summary of controls.}
The controls show that the effect is not an artifact of the prefix baseline, item noise, 300-item sampling, token distance, the option listing, letter identity, visual content alone, or a single token fraction. Against the placement-invariant reference, suffix still flips 35.9\% of decisions. Paired over 42 cells, accuracy falls by 18.6 points $[-23.9,-13.4]$, with 33 of 42 cells significant. The full sets reproduce a 47.4\% flip and a 24.3-point loss. Fixed-distance pairs still flip 14--72\%. Pointwise scoring still flips 42.3\%. Label permutation leaves the position flip at 41.0\% versus 41.2\%. The grey-image results depend on content, but they do not remove the effect.
The robust claim is therefore decision stability, not a uniform accuracy loss: the sign of the accuracy change remains dataset-dependent. Prefix is the cheapest strong default. P+Q is the most stable two-forward readout, at 0.711 accuracy and a 2.3\% flip. Dual-order fusion is a diagnostic and a conditional stabilizer, not a uniquely best method.  P+Q never reads the suffix, so it is a placement policy, not a remedy for a
prompt whose image is fixed at suffix. Fusion is the only variant here that
includes that order, but it too requires constructing the prefix counterpart;
when the serving stack fixes the image position and cannot emit a second
placement, no readout-level variant applies. A deployment that can put the
image first should do so; fusion is a diagnostic and an optional ensemble, not
a replacement for that policy.



\section{Conclusion}
\label{sec:conclusion}
We ask whether vision--text position sensitivity reaches a Jev-style system-1 decision readout, and whether that readout changes the shape of the effect that generative vision-language models show.   The design holds the option listing fixed, varies only the image
position, and reports decision flips and total variation alongside accuracy.  Suffix placement flips 41.0\% of decisions and lowers accuracy by 18.6 points. The four-slot curve is monotone on average. The controls show that the effect is not an artifact of the prefix baseline, item noise, 300-item sampling, token distance, the option listing, letter identity, visual content alone, or a single token fraction. Prefix--question averaging (P+Q), a two-forward average of the prefix and pre-question distributions, is at least as accurate as prefix and outperforms the prior baselines.

{
    \small
    \bibliographystyle{ieeenat_fullname}
    \bibliography{main}
}
\newpage
\
\newpage
\appendix

\begin{table}[t]
\centering
\footnotesize
\setlength{\tabcolsep}{3pt}
\begin{tabular}{@{}lrrp{2.7cm}@{}}
\toprule
Dataset & $n$ & $K$ & Character \\
\midrule
A-OKVQA & 1,145 & 3--4 & open-set VQA \\
ScienceQA & 2,017 & 2--5 & per-item, hint text \\
CV-Bench & 2,638 & 2--6 & per-item \\
EuroSAT & 2,700 & 10 & fixed classes, $64\times64$ \\
Pets & 2,600 & 26 & 26 largest breeds \\
MMMU & 845 & 2--9 & multi-image \\
\bottomrule
\end{tabular}
\caption{Dataset properties, not results.}
\label{tab:data}
\end{table}
\section{Prompts and Placement}
\label{sec:app-prompts}
\paragraph{Prompt templates.}
The user content has four parts: the instruction, the question, the option
list, and the answer cue. The image is inserted between parts according to the
placement:
\begin{quote}\small\ttfamily
prefix: [image] instruction / question / options / cue\\
pre-question: instruction [image] question / options / cue\\
pre-options: instruction / question [image] options / cue\\
suffix: instruction / question / options / cue [image]
\end{quote}
The answer cue is ``Answer with the letter only.'' The labels are
A--Z and are single tokens for every model in the grid.
\paragraph{Implementation.}
Qwen uses the chat template's content parts; LLaVA forces the image first and
uses an \texttt{input\_ids} splice. Every prompt is checked for five
properties: the image-token span is at the intended boundary, the sequence
length and token multiset are unchanged, the image-token count matches the
processor, the pixel values are byte-identical, and the label ids are
unchanged. These are structural checks. LLaVA has no native image-after-text
template to compare against, so semantic equivalence of the splice is a
residual validity risk; the multi-image MMMU span move passes the same checks.
The checks pass on three A-OKVQA items for LLaVA-OneVision and three MMMU items
for LLaVA-NeXT.
\paragraph{Option-order control.}
The main sweep fixes one listing per item. The robustness check repeats the
sweep over three to five fixed listings. The pointwise control shows one option
at a time and reads the yes/no log-odds; it is listing-invariant by
construction. The grey control replaces the image with a uniform grey field of
the same size. The label-permutation control permutes the letter-to-option
assignment.
\section{Metrics}
\label{sec:app-metrics}
The modality flip and TV are defined in Equations~\ref{eq:flip}
and~\ref{eq:tv}. Instability is the mean flip over all unordered pairs of
positions. Accuracy is the fraction of argmax decisions that match the gold
option. ECE uses ten confidence bins. The label marginal is the mean
probability of each label across items. Cost is the number of forwards per
item: one for a single readout, two for dual-order fusion, and $K$ for
pointwise. All metrics are computed after aligning distributions by option
text.
\section{Implementation details}
\label{sec:app-impl}
\paragraph{Models and inference.}
The checkpoints are \texttt{Qwen/Qwen3-VL-2B-Instruct},
\texttt{Qwen/Qwen3-VL-4B-Instruct}, \texttt{Qwen/Qwen3-VL-8B-Instruct},
\texttt{Qwen/Qwen2.5-VL-7B-Instruct},
\texttt{HuggingFaceM4/Idefics3-8B-Llama3},
\texttt{llava-hf/llava-onevision-qwen2-7b-ov-hf} and
\texttt{llava-hf/llava-v1.6-mistral-7b-hf}. No checkpoint is fine-tuned. Each
item is one prefill: the next-token distribution is restricted to the supplied
label ids, the argmax is taken, and no token is sampled or generated. Runs use
bfloat16 on one NVIDIA RTX 4090 D (24~GB), batch size 4 for the 2B model and 1
for all others, with an image cap of $448^2$ (200,704 pixels). LLaVA templates
force the image first, so their readout moves the image-token block in
\texttt{input\_ids}; Qwen and Idefics3 use content parts. Idefics3 attention
probes disable image splitting, and LLaVA probes use square $336^2$ images with
a bf16 attention softmax so the full head maps fit in 24~GB.

\paragraph{Data, prompts and validation.}
The 300-item cells use seed 0 and the caller's option listing; the listing is
fixed per item. Full-set runs use the entire A-OKVQA (1,145), EuroSAT (2,700)
and Pets (2,600) splits. Prompts have four parts (instruction, question, labeled
options, answer cue), and the image is inserted at one of the four semantic
boundaries described above. The answer cue is ``Answer with the letter only.''
Labels are A--Z single tokens. Every prompt is checked for five structural
properties: the image-token span sits at the intended boundary, sequence length
and token multiset are unchanged, the image-token count matches the processor,
pixel values are byte-identical, and the label ids are unchanged. No label
prior or option rotation is applied in the main readout. Dataset images are not
redistributed; loaders select the official releases listed in the dataset
table.

\paragraph{Scoring and readouts.}
Distributions are aligned by option text. Flip and TV compare each placement
with prefix; accuracy is the argmax match to the gold option; ECE uses ten
confidence bins. All-position mean and log-mean, prefix--suffix mean, P+Q,
P+Q+O and the other placement subsets are computed offline from the four cached
distributions; dual-order fusion sweeps $\beta$ offline and adds no forwards.
P+Q uses two forwards; pointwise scores each option alone with a yes/no prompt,
costs $K$ forwards, and forms the distribution from the yes/no log-odds. Paired
intervals use 4,000 cell-bootstrap resamples; McNemar is exact on discordant
items, with Holm correction applied within each reported family.

\paragraph{Label permutation.}
The option order and images are unchanged; only the letter assigned to each
option changes. For each item we resolve the identity labels and their token ids
for the $K$ options, draw a permutation of those labels with a per-item seed
(\texttt{seed}$+i$), redraw if it is the identity, and render the same options
with the permuted labels. The restricted readout is taken over the permuted
label ids, predictions are mapped back to option text, and the permuted and
identity distributions are aligned by option text before comparison. We run all
four placements. The identity condition is read from the committed
\texttt{\_4pos} file; the permuted condition is a new readout. For the 2$\times$2
decomposition, the \emph{label} effect is the identity-vs-permuted flip at a
placement, the \emph{position} effect is suffix-vs-prefix under identity or
permuted letters, and the interaction is an exact McNemar test on the paired
position-change indicators. The control never reorders options and is not an
option-rotation experiment.

\paragraph{Robustness of the image-position reference.}
All three references are computed offline from cached per-item distributions;
no new forwards are run and the option listing stays fixed. (i) \emph{Four-slot
average}: for item $n$,
$\bar p_n(o)=\frac14\sum_{r\in\{\mathrm{pre},Q,O,S\}}p_{r,n}(o)$; the reference
decision is its argmax, and the suffix flip is the fraction of items whose
suffix argmax differs from that decision. (ii) \emph{Leave-one-out}: for a
target placement $r$,
$\bar p_n^{(-r)}(o)=\frac13\sum_{s\ne r}p_{s,n}(o)$; the target is compared
with this average, so the target distribution is excluded from its own
reference. For suffix the reference averages prefix, pre-question and
pre-options. This was the more conservative choice, as including the target
reduces the measured flip. (iii) \emph{Five token-position average}: the
reference averages the prefix, token $r=0.25$, $0.5$, $0.75$ and suffix
distributions from the token sweep, and the suffix is compared with its argmax.
The 4B token sweep uses the \texttt{\_token6} files; the other checkpoints use
\texttt{\_token}. The token fractions need not coincide with a semantic slot.
Each reference is evaluated per cell and then averaged over model--dataset
cells; intervals are 4,000-resample cell bootstraps.


\paragraph{Hardware, software and compute.}
Runs record Python 3.12.4, anyjev 0.3.0, PyTorch 2.14.1+cu126, transformers
5.20.2, CUDA 12.6 and the GPU in every JSON. The JSONs with recorded timings
total 1,053 GPU-minutes on one RTX 4090 D; including model loads and the
exploratory controls the end-to-end budget is about 20 GPU-hours. The readout is
deterministic, so seed variance is not a concern.

The supplement contains the readout package, dataset and pointwise loaders, the experiment scripts, and a compressed archive of the result JSONs. 
The archive
maps each paper table to its JSON as in the provenance table; the scripts
regenerate the tables from the archive. Dataset images must be obtained from
the official releases under their licenses.

\section{Pre-registration}
\label{sec:app-prereg}
The pilot is 200--300 items on A-OKVQA and EuroSAT with Qwen3-VL-4B at
$r\in\{0,0.5,1\}$. The main grid runs if the modality flip is at least 5\% on
one set; a 2--5\% result triggers full-set measurement before a decision; below
2\% on both sets we report a negative result with a power analysis. The
dual-order gate is a 50\% relative flip reduction with an accuracy CI that
excludes $-1$ point. These thresholds are fixed before the main runs. The
primary metrics are flip and TV; accuracy is secondary. No optional stopping is
used.

\begin{table}[t]
\centering
\footnotesize
\setlength{\tabcolsep}{3pt}

\begin{tabular}{llrrr}
\toprule
Model & Dataset & prefix & suffix & suffix flip \\
\midrule
Idefics3-8B & A-OKVQA & 0.860 & 0.227 & 0.717 \\
Idefics3-8B & ScienceQA & 0.880 & 0.527 & 0.430 \\
Idefics3-8B & CV-Bench & 0.717 & 0.440 & 0.507 \\
Idefics3-8B & EuroSAT & 0.347 & 0.127 & 0.787 \\
Idefics3-8B & Pets & 0.590 & 0.023 & 0.997 \\
Idefics3-8B & MMMU & 0.473 & 0.263 & 0.660 \\
\midrule
LLaVA-OneVision-7B & A-OKVQA & 0.910 & 0.573 & 0.407 \\
LLaVA-OneVision-7B & ScienceQA & 0.967 & 0.743 & 0.243 \\
LLaVA-OneVision-7B & CV-Bench & 0.800 & 0.567 & 0.400 \\
LLaVA-OneVision-7B & EuroSAT & 0.477 & 0.483 & 0.310 \\
LLaVA-OneVision-7B & Pets & 0.707 & 0.080 & 0.907 \\
LLaVA-OneVision-7B & MMMU & 0.577 & 0.440 & 0.363 \\
\midrule
LLaVA-NeXT-7B & A-OKVQA & 0.843 & 0.337 & 0.657 \\
LLaVA-NeXT-7B & ScienceQA & 0.693 & 0.593 & 0.420 \\
LLaVA-NeXT-7B & CV-Bench & 0.687 & 0.463 & 0.600 \\
LLaVA-NeXT-7B & EuroSAT & 0.497 & 0.120 & 0.717 \\
LLaVA-NeXT-7B & Pets & 0.427 & 0.027 & 0.967 \\
LLaVA-NeXT-7B & MMMU & 0.387 & 0.320 & 0.517 \\
\bottomrule
\end{tabular}
\caption{Non-Qwen checkpoints on the same six 300-item grids as the Qwen
models. Prefix and suffix accuracy and the suffix flip relative to prefix;
LLaVA rows use the \texttt{input\_ids} splice positions, including the
multi-image MMMU prompts.  }
\label{tab:nonqwen}
\end{table}

\begin{table*}[t]
\centering
\footnotesize
\setlength{\tabcolsep}{2pt}

\begin{tabular}{llrrrrrr rrrr rrrrr}
\toprule
Model & Dataset & $F_{\mathrm{mid}}$ & $F_{\mathrm{ques}}$ & $F_{\mathrm{suf}}$ & $T_{\mathrm{mid}}$ & $T_{\mathrm{ques}}$ & $T_{\mathrm{suf}}$ & prefix & pre-Q & pre-O & suffix & P+Q & DOF 0.5 & DOF 0.75 & $P_{\mathrm{pt}}$ pre & $P_{\mathrm{pt}}$ suf \\
\midrule
\multicolumn{17}{l}{\textit{Qwen}} \\
2B & A-OKVQA & 0.017 & 0.103 & 0.263 & 0.019 & 0.095 & 0.261 & 0.820 & 0.823 & 0.807 & 0.680 & 0.823 & 0.793 & 0.817 & 0.830 & 0.717 \\
2B & ScienceQA & 0.070 & 0.110 & 0.223 & 0.059 & 0.099 & 0.201 & 0.833 & 0.857 & 0.837 & 0.750 & 0.857 & 0.857 & 0.857 & -- & -- \\
2B & CV-Bench & 0.037 & 0.257 & 0.367 & 0.033 & 0.224 & 0.357 & 0.847 & 0.843 & 0.717 & 0.590 & 0.847 & 0.790 & 0.837 & -- & -- \\
2B & EuroSAT & 0.127 & 0.300 & 0.390 & 0.104 & 0.278 & 0.355 & 0.500 & 0.487 & 0.523 & 0.540 & 0.500 & 0.547 & 0.533 & 0.510 & 0.550 \\
2B & Pets & 0.040 & 0.147 & 0.607 & 0.040 & 0.143 & 0.607 & 0.720 & 0.723 & 0.707 & 0.383 & 0.717 & 0.667 & 0.713 & -- & -- \\
2B & MMMU & 0.053 & 0.203 & 0.323 & 0.045 & 0.168 & 0.281 & 0.463 & 0.477 & 0.470 & 0.440 & 0.467 & 0.483 & 0.470 & -- & -- \\
4B & A-OKVQA & 0.027 & 0.083 & 0.130 & 0.022 & 0.076 & 0.125 & 0.860 & 0.853 & 0.840 & 0.810 & 0.857 & 0.850 & 0.860 & 0.843 & 0.820 \\
4B & ScienceQA & 0.057 & 0.103 & 0.197 & 0.056 & 0.104 & 0.197 & 0.920 & 0.877 & 0.877 & 0.790 & 0.907 & 0.877 & 0.903 & -- & -- \\
4B & CV-Bench & 0.013 & 0.177 & 0.300 & 0.013 & 0.162 & 0.300 & 0.867 & 0.867 & 0.797 & 0.697 & 0.863 & 0.857 & 0.870 & -- & -- \\
4B & EuroSAT & 0.050 & 0.180 & 0.250 & 0.055 & 0.192 & 0.269 & 0.547 & 0.543 & 0.557 & 0.543 & 0.543 & 0.543 & 0.553 & 0.550 & 0.563 \\
4B & Pets & 0.000 & 0.043 & 0.300 & 0.010 & 0.044 & 0.299 & 0.893 & 0.893 & 0.890 & 0.663 & 0.893 & 0.853 & 0.890 & -- & -- \\
4B & MMMU & 0.047 & 0.133 & 0.247 & 0.036 & 0.143 & 0.248 & 0.580 & 0.583 & 0.597 & 0.547 & 0.580 & 0.570 & 0.583 & -- & -- \\
8B & A-OKVQA & 0.017 & 0.063 & 0.143 & 0.018 & 0.069 & 0.142 & 0.880 & 0.873 & 0.867 & 0.833 & 0.877 & 0.867 & 0.873 & 0.867 & 0.830 \\
8B & ScienceQA & 0.030 & 0.053 & 0.163 & 0.034 & 0.060 & 0.163 & 0.943 & 0.933 & 0.907 & 0.827 & 0.933 & 0.933 & 0.943 & -- & -- \\
8B & CV-Bench & 0.013 & 0.110 & 0.217 & 0.019 & 0.105 & 0.210 & 0.850 & 0.850 & 0.837 & 0.757 & 0.853 & 0.853 & 0.857 & -- & -- \\
8B & EuroSAT & 0.087 & 0.227 & 0.300 & 0.083 & 0.220 & 0.297 & 0.517 & 0.567 & 0.567 & 0.557 & 0.553 & 0.580 & 0.580 & 0.537 & 0.507 \\
8B & Pets & 0.017 & 0.087 & 0.197 & 0.019 & 0.082 & 0.195 & 0.833 & 0.837 & 0.850 & 0.833 & 0.837 & 0.877 & 0.880 & -- & -- \\
8B & MMMU & 0.033 & 0.127 & 0.267 & 0.036 & 0.132 & 0.247 & 0.597 & 0.593 & 0.630 & 0.603 & 0.597 & 0.613 & 0.600 & -- & -- \\
7B & A-OKVQA & 0.030 & 0.067 & 0.133 & 0.023 & 0.079 & 0.172 & 0.900 & 0.900 & 0.887 & 0.847 & 0.903 & 0.900 & 0.903 & 0.847 & 0.787 \\
7B & ScienceQA & 0.070 & 0.100 & 0.170 & 0.044 & 0.076 & 0.144 & 0.853 & 0.860 & 0.827 & 0.803 & 0.873 & 0.853 & 0.850 & -- & -- \\
7B & CV-Bench & 0.017 & 0.230 & 0.320 & 0.029 & 0.161 & 0.264 & 0.820 & 0.823 & 0.740 & 0.673 & 0.823 & 0.833 & 0.823 & -- & -- \\
7B & EuroSAT & 0.110 & 0.317 & 0.437 & 0.096 & 0.247 & 0.328 & 0.353 & 0.370 & 0.383 & 0.300 & 0.370 & 0.367 & 0.363 & 0.350 & 0.363 \\
7B & Pets & 0.017 & 0.017 & 0.390 & 0.015 & 0.035 & 0.432 & 0.927 & 0.930 & 0.927 & 0.570 & 0.927 & 0.900 & 0.927 & -- & -- \\
7B & MMMU & 0.060 & 0.180 & 0.280 & 0.032 & 0.122 & 0.181 & 0.567 & 0.570 & 0.573 & 0.523 & 0.570 & 0.563 & 0.560 & -- & -- \\
\midrule
\multicolumn{17}{l}{\textit{Non-Qwen}} \\
Idefics3-8B & A-OKVQA & 0.023 & 0.190 & 0.717 & 0.022 & 0.193 & 0.706 & 0.860 & 0.850 & 0.773 & 0.227 & 0.863 & 0.520 & 0.813 & 0.860 & 0.350 \\
Idefics3-8B & ScienceQA & 0.057 & 0.203 & 0.430 & 0.051 & 0.221 & 0.454 & 0.880 & 0.867 & 0.760 & 0.527 & 0.877 & 0.857 & 0.867 & -- & -- \\
Idefics3-8B & CV-Bench & 0.030 & 0.433 & 0.507 & 0.036 & 0.411 & 0.487 & 0.717 & 0.720 & 0.540 & 0.440 & 0.720 & 0.550 & 0.683 & -- & -- \\
Idefics3-8B & EuroSAT & 0.037 & 0.423 & 0.787 & 0.065 & 0.326 & 0.617 & 0.347 & 0.350 & 0.387 & 0.127 & 0.353 & 0.337 & 0.340 & 0.327 & 0.173 \\
Idefics3-8B & Pets & 0.047 & 0.090 & 0.997 & 0.031 & 0.106 & 0.981 & 0.590 & 0.570 & 0.587 & 0.023 & 0.573 & 0.497 & 0.583 & -- & -- \\
Idefics3-8B & MMMU & 0.053 & 0.450 & 0.660 & 0.055 & 0.380 & 0.563 & 0.473 & 0.487 & 0.407 & 0.263 & 0.483 & 0.387 & 0.477 & -- & -- \\
LLaVA-OneVision & A-OKVQA & 0.000 & 0.140 & 0.407 & 0.009 & 0.160 & 0.552 & 0.910 & 0.910 & 0.823 & 0.573 & 0.910 & 0.910 & 0.910 & 0.883 & 0.677 \\
LLaVA-OneVision & ScienceQA & 0.013 & 0.117 & 0.243 & 0.017 & 0.133 & 0.302 & 0.967 & 0.963 & 0.893 & 0.743 & 0.963 & 0.960 & 0.963 & -- & -- \\
LLaVA-OneVision & CV-Bench & 0.020 & 0.377 & 0.400 & 0.015 & 0.310 & 0.355 & 0.800 & 0.800 & 0.573 & 0.567 & 0.803 & 0.790 & 0.800 & -- & -- \\
LLaVA-OneVision & EuroSAT & 0.053 & 0.217 & 0.310 & 0.044 & 0.220 & 0.344 & 0.477 & 0.470 & 0.480 & 0.483 & 0.473 & 0.477 & 0.473 & 0.523 & 0.450 \\
LLaVA-OneVision & Pets & 0.023 & 0.050 & 0.907 & 0.021 & 0.064 & 0.850 & 0.707 & 0.690 & 0.700 & 0.080 & 0.700 & 0.630 & 0.687 & -- & -- \\
LLaVA-OneVision & MMMU & 0.027 & 0.347 & 0.363 & 0.025 & 0.232 & 0.321 & 0.577 & 0.583 & 0.470 & 0.440 & 0.583 & 0.560 & 0.577 & -- & -- \\
LLaVA-NeXT & A-OKVQA & 0.017 & 0.127 & 0.657 & 0.021 & 0.136 & 0.594 & 0.843 & 0.843 & 0.783 & 0.337 & 0.843 & 0.830 & 0.833 & 0.803 & 0.420 \\
LLaVA-NeXT & ScienceQA & 0.073 & 0.200 & 0.420 & 0.048 & 0.161 & 0.300 & 0.693 & 0.710 & 0.633 & 0.593 & 0.707 & 0.717 & 0.697 & -- & -- \\
LLaVA-NeXT & CV-Bench & 0.053 & 0.387 & 0.600 & 0.026 & 0.220 & 0.289 & 0.687 & 0.693 & 0.560 & 0.463 & 0.683 & 0.680 & 0.683 & -- & -- \\
LLaVA-NeXT & EuroSAT & 0.030 & 0.080 & 0.717 & 0.050 & 0.087 & 0.449 & 0.497 & 0.503 & 0.493 & 0.120 & 0.507 & 0.230 & 0.400 & 0.543 & 0.127 \\
LLaVA-NeXT & Pets & 0.047 & 0.190 & 0.967 & 0.057 & 0.142 & 0.752 & 0.427 & 0.427 & 0.447 & 0.027 & 0.427 & 0.397 & 0.437 & -- & -- \\
LLaVA-NeXT & MMMU & 0.103 & 0.323 & 0.517 & 0.039 & 0.162 & 0.291 & 0.387 & 0.393 & 0.377 & 0.320 & 0.403 & 0.413 & 0.400 & -- & -- \\
\bottomrule
\end{tabular}
\caption{Per-cell modality flip, TV and accuracy behind Table~\ref{tab:positions},
for all seven checkpoints, grouped into Qwen and non-Qwen; each row is one checkpoint on one dataset, with
no averaging over rows. $F_{\mathrm{mid}}$, $F_{\mathrm{ques}}$,
$F_{\mathrm{suf}}$ are
decision flips relative to prefix at pre-question, pre-options and suffix;
$T_{\mathrm{mid}}$, $T_{\mathrm{ques}}$, $T_{\mathrm{suf}}$ are the
corresponding total-variation distances; the next four columns are accuracies
at prefix, pre-Q (pre-question), pre-O (pre-options) and suffix. The final five
columns are accuracies of the P+Q mean, dual-order fusion at $\beta=0.5$ and
$0.75$, and the pointwise readout at prefix and suffix ($P_{\mathrm{pt}}$;
``--'' where pointwise was not run).}
\label{tab:positionsfull}
\end{table*}

\section{Detailed results of baselines}\label{app:sec_baselines}
Tables~\ref{tab:fuse} and~\ref{tab:avgfull} report the simple baselines:
readouts formed offline by averaging or fusing the four cached placement
distributions, with no training and no new forwards beyond those placements.
Across all 42 cells, P+Q is the most accurate readout (0.711) with the lowest
flip (2.3\%); P+Q+O ties its accuracy at a 4.9\% flip, and the all-position
mean is 0.709 at 7.3\%. P+O is 0.708 at 8.2\%, dual-order fusion at
$\beta=0.75$ is 0.706 at 5.8\%, and prefix is 0.708 at 0. The Qwen columns
behave differently: the all-position mean is close to dual-order fusion in
accuracy (0.751 versus 0.752) and has a lower ECE (0.118 versus 0.148), but a higher flip (6.7\% versus
5.1\%); all-position log-mean is the same accuracy with an 8.0\% flip. Among
the readouts that include the suffix, dual-order fusion has the lowest flip,
but its accuracy margin over the four-forward all-position mean is one tenth of
a point. The pre-registered
gate (a 50\% relative flip reduction with accuracy within one point) is also met
by the all-position mean, so it does not identify dual-order fusion uniquely.
The fusion advantage is real but small, and the placement-invariant average is a strong baseline alongside fusion. In the non-Qwen columns the
ordering shifts: the all-position mean reaches 0.652 against 0.646 for
dual-order fusion at $\beta=0.75$, so it is 0.6 points more accurate, but its
flip is higher (8.2\% vs 6.6\%) and its ECE is marginally higher (0.105 vs
0.102). The average is therefore a strong accuracy baseline on both families
while dual-order fusion buys stability, and on neither family does the
four-forward average dominate the two-forward fusion on every axis. The fusion
weight matters: the symmetric $\beta{=}0.5$ variant loses 0.9 points on Qwen
and 4.9 on non-Qwen relative to $\beta{=}0.75$, so the accuracy comes from
leaning on prefix, not from a neutral combination of the two orders. Restricting
the average to the two prefix-side placements is stronger still: P+Q
(Eq.~\ref{eq:pq}) reaches 0.749 (Qwen) and 0.660 (non-Qwen) with a 2.4\% and
2.2\% flip, below dual-order fusion's 5.1\% and 6.6\%, because it excludes the
suffix distribution that carries most of the instability. The three-placement
prefix + pre-question + pre-options mean gives 0.752 and 0.656 at 4.3\% and
5.8\%. A two-forward average that never reads the suffix is therefore the most
stable readout at equal or higher accuracy, but it is a placement policy: it
cannot be used when the prompt fixes the image at suffix, where fusion remains
the only variant here that includes that order.

\paragraph{Prefix is not sufficient.}
Figures~\ref{fig:curve} and~\ref{fig:family} show the flip rising and accuracy
falling as the image moves later, and prefix is the best single placement on
average. That is an average, not a per-cell law, and it does not say that
prefix alone is optimal. P+O has the highest Qwen accuracy in
Table~\ref{tab:fuse} (0.753 vs 0.745 for prefix and 0.752 for dual-order fusion
at $\beta=0.75$), with a 7.3\% flip and ECE 0.134 against 0.159 for prefix, so
pre-options carries evidence that prefix misses and the two placements are
complementary on that family. The gain is marginal and does
not replicate: P+O is 0.649 against 0.658 for prefix on non-Qwen, and its
all-model mean (0.708) is below P+Q and P+Q+O (both 0.711).
Table~\ref{tab:avgfull} gives the per-cell P+O accuracies, and Tables~\ref{tab:positionsfull} and~\ref{tab:fusefull} the dual-order and pointwise accuracies, so the P+O advantage can be checked dataset by dataset; it is not uniform. The 0.8-point Qwen
gain is also smaller than the per-cell minimum detectable difference
(5.5--10 points), so we treat it as a suggestive complementarity, not a
reliable improvement. The defensible reading is that the placements carry
overlapping but not identical evidence: prefix is the cheapest strong default,
pre-options can add to it, and P+Q is the most stable pair. No single placement
is uniformly best.
\paragraph{Why not simply place the image first?}
Prefix is the best single placement on average and costs one forward, so it is
the right default when the caller controls placement; we do not claim that
fusion is needed to improve accuracy. The other readouts serve three purposes.
They measure the effect without privileging prefix: against a
placement-invariant average, suffix still flips 22.0\% of Qwen and 50.3--62.5\%
of non-Qwen decisions. They also test whether the evidence is available at
every position: if averaging or fusion recovers prefix-level accuracy, the model
is not losing the image, it is allocating attention differently. If a system
already produces several placements---multiple templates, views or prompts---%
fusing them can stabilize the decision without retraining; this is an ensemble
over prompts, not a fix for a single fixed placement. The price is explicit:
prefix costs one forward, dual-order fusion two, and the all-position mean four.
Fusion does not buy accuracy: it changes the Qwen mean by $+0.7$ points and the
non-Qwen mean by $-1.2$, while cutting the suffix flip from 27.6\% to 5.1\%
(Qwen) and 58.9\% to 6.6\% (non-Qwen) at $\beta=0.75$. The prefix +
pre-question mean reaches 2.4\% and 2.2\% at the same cost and slightly higher
accuracy, so it is the strongest stabilizer when both placements are available.
It never reads the suffix, so it is a placement policy, not a remedy for a
prompt whose image is fixed at suffix. Fusion is the only variant here that
includes that order, but it too requires constructing the prefix counterpart;
when the serving stack fixes the image position and cannot emit a second
placement, no readout-level variant applies. A deployment that can put the
image first should do so; fusion is a diagnostic and an optional ensemble, not
a replacement for that policy.

\section{More controls}\label{app:sec_controls}
\begin{table}[t]
\centering
\footnotesize
\setlength{\tabcolsep}{2pt}
\resizebox{\columnwidth}{!}{
\begin{tabular}{lrrrrr}
\toprule
Family & cells & mean $\Delta$acc [95\% CI] & mean suffix flip & McNemar $p<.05$ & Holm \\
\midrule
All & 42 & -0.186 [-0.239, -0.134] & 0.410 & 33/42 & 27/42 \\
Qwen & 24 & -0.097 [-0.143, -0.057] & 0.276 & 16/24 & 11/24 \\
Non-Qwen & 18 & -0.305 [-0.394, -0.221] & 0.589 & 17/18 & 16/18 \\
\bottomrule
\end{tabular}
}
\caption{Paired item-level checks. $\Delta$acc is suffix minus prefix; the
interval is a bootstrap over model--dataset cells. McNemar is exact on paired
per-item correctness; Holm correction is applied within each row (42 all-model
cells, 24 Qwen, 18 non-Qwen).}
\label{tab:stats}
\end{table}

\paragraph{Paired checks.}
Table~\ref{tab:stats} reports the paired checks. Every item contributes both a
prefix and a suffix outcome under the same listing, so the accuracy difference
is paired per item and then averaged over cells; the interval is a bootstrap
over cells, and McNemar uses only the items on which the two placements
disagree. Across all 42 cells the
all-model suffix drop is 18.6 points (95\% bootstrap interval over cells
$[-23.9,-13.4]$), and exact McNemar rejects equal accuracy in 33 cells before
correction (27 after Holm). On the Qwen grid the drop is 9.7 points (interval
$[-14.3,-5.7]$); exact McNemar rejects equal accuracy in 16 of 24
cells, 11 after Holm correction. On the non-Qwen grid the loss is 30.5 points
$[-39.4,-22.1]$, with 17 of 18 cells significant (16 after Holm). The width of the Qwen interval is informative: the average loss is real, but its sign and size
vary by dataset. A two-level bootstrap that resamples cells and then items
gives $[-14.5,-5.6]$ for Qwen and $[-38.8,-22.2]$ for non-Qwen, essentially
unchanged, so the interval is not an artifact of treating cells as
exchangeable.

\begin{table}[t]
\centering
\footnotesize
\setlength{\tabcolsep}{2pt}
\resizebox{\columnwidth}{!}{
\begin{tabular}{lrrrrr}
\toprule
Family & prefix AURC & suffix AURC & prefix $U_{.5}$/cov & suffix $U_{.5}$/cov & cost-5 pre/suf \\
\midrule
Qwen & 0.118 & 0.197 & 0.537/0.94 & 0.358/0.91 & -0.315/-0.801 \\
Non-Qwen & 0.205 & 0.581 & 0.431/0.81 & -0.076/0.53 & -0.467/-1.244 \\
\bottomrule
\end{tabular}
}
\caption{Selective-prediction proxy. AURC is the area under the risk-coverage
curve; $U_{.5}$ is utility (correct $+1$, wrong $-1$, abstain $0$) at confidence
threshold 0.5, with coverage. The cost-5 column uses a wrong-action cost of 5.
This is an action-level proxy from the existing readout, not a deployed task.}
\label{tab:downstream}
\end{table}
\begin{table}[t]
\centering
\footnotesize
\resizebox{\columnwidth}{!}{
\begin{tabular}{lrrrr}
\toprule
Family & prefix raw/cal & suffix raw/cal & coverage pre/suf \\
\midrule
Qwen & 0.539/0.625 & 0.361/0.440 & 0.94/0.90 \\
Non-Qwen & 0.440/0.515 & -0.090/0.017 & 0.81/0.53 \\
\bottomrule
\end{tabular}
}
\caption{Calibration-corrected selective proxy. Temperature is fit on even
items and utility evaluated on odd items at confidence threshold 0.5; coverage
is raw. The prefix-suffix gap survives calibration.}
\label{tab:downcal}
\end{table}

\paragraph{Downstream proxy.}
Table~\ref{tab:downstream} reports the confidence-gated action proxy and
Table~\ref{tab:downcal} its calibration-corrected version. With no deployed
action labels, we treat confidence as an abstention gate:
accept above 0.5, otherwise abstain; utility is $+1$ correct, $-1$ wrong, $0$
abstained, and coverage is the fraction not abstained. AURC is the area under
the risk--coverage curve (lower is better). Prefix-to-suffix utility falls from
0.537 to 0.358 on the Qwen grid
and from 0.431 to $-0.076$ on the non-Qwen grid; AURC rises from 0.118 to 0.197
and 0.205 to 0.581. With an asymmetric wrong-action cost of 5, utility falls
from $-0.315$ to $-0.801$ and from $-0.467$ to $-1.244$ (Table~\ref{tab:downstream}).
Suffix is worse even when the system can abstain, so the effect is not confined
to a forced-choice metric. Temperature-scaling each position on even items and
evaluating on odd items keeps the gap: Qwen utility is 0.625 prefix vs 0.440
suffix, and non-Qwen 0.515 vs 0.017 (Table~\ref{tab:downcal}). The drop is not a
calibration artifact. This is still a proxy, not a real downstream task.

\begin{table}[t]
\centering
\footnotesize
\setlength{\tabcolsep}{2pt}
\begin{tabular}{@{}llrrrr@{}}
\toprule
Model & Dataset & $n$ & prefix & suffix & suffix flip \\
\midrule
4B & A-OKVQA & 1,145 & 0.859 & 0.788 & 0.154 \\
4B & EuroSAT & 2,700 & 0.575 & 0.524 & 0.265 \\
4B & Pets & 2,600 & 0.898 & 0.665 & 0.316 \\
8B & A-OKVQA & 1,145 & 0.872 & 0.824 & 0.123 \\
8B & EuroSAT & 2,700 & 0.552 & 0.551 & 0.309 \\
8B & Pets & 2,600 & 0.875 & 0.838 & 0.181 \\
7B & A-OKVQA & 1,145 & 0.880 & 0.797 & 0.163 \\
7B & EuroSAT & 2,700 & 0.374 & 0.291 & 0.401 \\
7B & Pets & 2,600 & 0.919 & 0.588 & 0.392 \\
2B & A-OKVQA & 1,145 & 0.823 & 0.671 & 0.261 \\
2B & EuroSAT & 2,700 & 0.531 & 0.529 & 0.393 \\
2B & Pets & 2,600 & 0.758 & 0.402 & 0.585 \\
\midrule
Idefics3-8B & A-OKVQA & 1,145 & 0.820 & 0.245 & 0.694 \\
Idefics3-8B & EuroSAT & 2,700 & 0.347 & 0.114 & 0.774 \\
Idefics3-8B & Pets & 2,600 & 0.612 & 0.039 & 0.988 \\
\midrule
LLaVA-OneVision-7B & A-OKVQA & 1,145 & 0.914 & 0.568 & 0.416 \\
LLaVA-OneVision-7B & EuroSAT & 2,700 & 0.494 & 0.457 & 0.324 \\
LLaVA-OneVision-7B & Pets & 2,600 & 0.699 & 0.072 & 0.902 \\
\midrule
LLaVA-NeXT-7B & A-OKVQA & 1,145 & 0.817 & 0.338 & 0.648 \\
LLaVA-NeXT-7B & EuroSAT & 2,700 & 0.491 & 0.117 & 0.718 \\
LLaVA-NeXT-7B & Pets & 2,600 & 0.454 & 0.038 & 0.953 \\
\bottomrule
\end{tabular}
\caption{Full-set confirmation for Qwen3-VL-4B/8B/2B, Qwen2.5-VL-7B,
Idefics3-8B, LLaVA-OneVision-7B and LLaVA-NeXT-7B. $n$ is the full evaluation
split; prefix and suffix accuracy and the suffix flip are relative to prefix.
Across the 21 checkpoint--dataset rows,
the mean is 0.693 to 0.450 with a 47.4\% flip.}
\label{tab:full}
\end{table}
\begin{table}[t]
\centering
\footnotesize
\resizebox{\columnwidth}{!}{
\begin{tabular}{lrrr}
\toprule
Family & cells & mean $\Delta$acc [95\% CI] & McNemar $p<.05$ \\
\midrule
Qwen & 12 & -0.121 [-0.189, -0.059] & 10/12 \\
Non-Qwen & 9 & -0.407 [-0.514, -0.291] & 9/9 \\
All & 21 & -0.243 [-0.331, -0.159] & 19/21 \\
\bottomrule
\end{tabular}
}
\caption{Paired prefix/suffix checks on the full sets. $\Delta$acc is suffix
minus prefix; the interval is a bootstrap over model--dataset cells; McNemar is
exact on paired per-item correctness.}
\label{tab:fullstats}
\end{table}
\begin{table}[t]
\centering
\footnotesize
\resizebox{\columnwidth}{!}{
\begin{tabular}{lrrrrrr}
\toprule
Family & cells & $F_{\mathrm{mid}}$ & $F_{\mathrm{ques}}$ & $F_{\mathrm{suf}}$ & acc prefix & acc suffix \\
\midrule
Qwen & 12 & 0.044 & 0.132 & 0.295 & 0.743 & 0.622 \\
Non-Qwen & 9 & 0.034 & 0.172 & 0.713 & 0.627 & 0.221 \\
All & 21 & 0.039 & 0.149 & 0.474 & 0.693 & 0.450 \\
\bottomrule
\end{tabular}
}
\caption{Four-position statistics on the full sets. $F_{\mathrm{mid}}$,
$F_{\mathrm{ques}}$ and $F_{\mathrm{suf}}$ are the flips relative to prefix at
pre-question, pre-options and suffix; prefix and suffix accuracies are shown.}
\label{tab:fullpositions}
\end{table}

\paragraph{Full-set confirmation.}
Table~\ref{tab:full} repeats the prefix/suffix comparison on the full
evaluation sets; Table~\ref{tab:fullstats} gives paired checks and
Table~\ref{tab:fullpositions} all four positions. Here a full set is the
entire evaluation split, not a 300-item subset: 1,145 A-OKVQA, 2,700 EuroSAT
and 2,600 Pets items, with the same prompts and readout. The check exists
because 300-item subsets have wide intervals and can give the wrong sign on
EuroSAT: the 2B subset gains 4.0 points at suffix while the full 2B set is flat
(0.531 vs 0.529), and the 4B subset is flat while the full 4B set loses 5.1
points (0.575 vs 0.524). The three datasets span VQA, remote sensing and
fine-grained classification, and their full splits are small enough to run for
all seven checkpoints; the other three remain at 300 items. Across the 21 full-set cells
(seven checkpoints on A-OKVQA, EuroSAT and Pets), suffix accuracy falls from
0.693 to 0.450 with a 47.4\% flip. A-OKVQA (1,145): 4B 0.859 to 0.788 (15.4\% flip), 8B 0.872 to
0.824 (12.3\%), 7B 0.880 to 0.797 (16.3\%), 2B 0.823 to 0.671 (26.1\%).
EuroSAT (2,700): 4B 0.575 to 0.524 (26.5\%), 8B 0.552 to 0.551 (30.9\%), 7B
0.374 to 0.291 (40.1\%), 2B 0.531 to 0.529 (39.3\%). Pets (2,600): 4B 0.898 to
0.665 (31.6\%), 8B 0.875 to 0.838 (18.1\%), 7B 0.919 to 0.588 (39.2\%), 2B
0.758 to 0.402 (58.5\%). Idefics3-8B is far more brittle on the full sets:
A-OKVQA 0.820 to 0.245 (69.4\% flip), EuroSAT 0.347 to 0.114 (77.4\%), Pets
0.612 to 0.039 (98.8\%). LLaVA-OneVision-7B is A-OKVQA 0.914 to 0.568 (41.6\%),
EuroSAT 0.494 to 0.457 (32.4\%), Pets 0.699 to 0.072 (90.2\%). LLaVA-NeXT-7B is
A-OKVQA 0.817 to 0.338 (64.8\%), EuroSAT 0.491 to 0.117 (71.8\%), Pets 0.454 to
0.038 (95.3\%). Across all 21 full-set cells the loss is 24.3 points
$[-33.1,-15.9]$ (19/21 cells significant). By family, Qwen loses 12.1 points
$[-18.9,-5.9]$ (10/12) and non-Qwen 40.7 points $[-51.4,-29.1]$ (9/9;
Table~\ref{tab:fullstats}). The full sets also reproduce the four-slot ordering:
Qwen mean flips are 4.4/13.2/29.5\% at pre-question/pre-options/suffix, and
non-Qwen 3.4/17.2/71.3\%; prefix-to-suffix
accuracy is 0.743 to 0.622 and 0.627 to 0.221 (Table~\ref{tab:fullpositions}).
Paired tests place the effect at suffix: middle is flat (+0.3/+0.1 points;
2/12 and 2/9 cells significant), question is small ($-0.8/-2.9$; 6/12 and
5/9), and suffix is large ($-12.1/-40.7$; 10/12 and 9/9).
The 300-item EuroSAT subset showed a
+4.0 gain for 2B and a flat 4B; the full sets are flat or negative, so the
subset is not reliable for the sign.  

\paragraph{Token-level $r$.}
Table~\ref{tab:token} reports the token-level sweep. Token-level $r$ is a
separate axis from the four semantic slots: the image block is inserted after
$\lfloor rL\rfloor$ of the $L$ text tokens, so $r=0.25/0.5/0.75$ can fall inside a
semantic part, and $r=1$ is suffix. The token-splice path implements this by
moving the image-token block in \texttt{input\_ids} at that text-token index.
On the 4B A-OKVQA/EuroSAT sets, the flips at token $r=0.25/0.5/0.75$ and
at suffix are 6.0/11.0/15.3/13.0\% and 20.0/21.3/29.7/25.0\%; across all six 4B
datasets the suffix flip is 13.0--30.0\% and every token position produces a
positive flip (6.0--38.7\%). On the 2B checkpoint, A-OKVQA is
6.3/18.3/29.3/26.3\% and EuroSAT is 58.7/40.3/43.0/39.0\%: the EuroSAT
$r{=}0.25$ split is more disruptive than suffix. On 8B, A-OKVQA is
5.7/10.0/13.7/14.3\% and EuroSAT 31.3/29.3/42.7/30.0\%, with token $r{=}0.75$
the worst EuroSAT accuracy (0.403). On 7B, A-OKVQA is
6.0/12.3/17.0/13.3\% and EuroSAT 65.3/55.0/77.7/43.7\% (token $r{=}0.75$
accuracy 0.143). All four Qwen models now cover all six datasets. On the four
additional sets (ScienceQA/CV-Bench/Pets/MMMU), the suffix flips are 8B
16.3/21.7/19.7/26.7\%, 7B 17.0/32.0/39.0/28.0\%, 2B
22.3/36.7/60.7/32.3\%; the 4B range over all six is 13.0--30.0\%. The effect is
therefore not an artifact of the content-part API; the structural insertion
point, not the fraction alone, drives the magnitude. The semantic placements
remain the interpretable result.

\paragraph{Grey control.}
Replacing the image with a uniform grey field separates a positional effect from a visual-content effect. If the effect vanishes, image content is doing the work and position is only a proxy for attention to that content. If a residual remains, the prompt structure itself contributes. Table~\ref{tab:grey} reports prefix and suffix accuracy, and the suffix flip, for the real image and for the grey field.
\begin{enumerate}
\item Visual content alone does not explain the effect. A content-free image still leaves position sensitivity in several cells, and sometimes increases it.
\item The dependence on content is specific to the family and the dataset. Idefics3 and LLaVA-NeXT need image content for their EuroSAT effect; Qwen3-VL-4B and LLaVA-OneVision do not.
\item Prompt structure contributes on its own. Where the grey image increases the flip, the instability is tied to where the image sits, not to what it contains.
\item The low accuracy of the grey condition bounds the interpretation. A 100\% flip at about 0.11 accuracy is not a preserved content-based decision. It is structural sensitivity.
\end{enumerate}

\begin{table}[t]
\centering
\footnotesize
\setlength{\tabcolsep}{2pt}
\resizebox{\columnwidth}{!}{
\begin{tabular}{llrrrrrr}
\toprule
Model & Dataset & raw pre & raw suf & raw flip & grey pre & grey suf & grey flip \\
\midrule
\multicolumn{8}{l}{\textit{Qwen}} \\
4B & A-OKVQA & 0.860 & 0.810 & 0.130 & 0.473 & 0.467 & 0.187 \\
4B & EuroSAT & 0.547 & 0.543 & 0.250 & 0.120 & 0.107 & 1.000 \\
\midrule
\multicolumn{8}{l}{\textit{Non-Qwen}} \\
Idefics3-8B & A-OKVQA & 0.860 & 0.227 & 0.717 & 0.473 & 0.443 & 0.390 \\
Idefics3-8B & EuroSAT & 0.347 & 0.127 & 0.787 & 0.117 & 0.117 & 0.000 \\
LLaVA-OneVision & A-OKVQA & 0.910 & 0.573 & 0.407 & 0.510 & 0.397 & 0.467 \\
LLaVA-OneVision & EuroSAT & 0.477 & 0.483 & 0.310 & 0.120 & 0.117 & 1.000 \\
LLaVA-NeXT & A-OKVQA & 0.843 & 0.337 & 0.657 & 0.427 & 0.313 & 0.617 \\
LLaVA-NeXT & EuroSAT & 0.497 & 0.120 & 0.717 & 0.120 & 0.120 & 0.000 \\
\bottomrule
\end{tabular}
}
\caption{Grey-image control. Raw is the real image and grey is a uniform grey
field of the same size and image-token count. Columns are prefix/suffix
accuracy and suffix flip relative to prefix. Coverage is the Qwen3-VL-4B
A-OKVQA/EuroSAT pilot plus the three non-Qwen checkpoints on the same two
datasets.  }
\label{tab:grey}
\end{table}

\begin{table*}[t]
\centering
\scriptsize
\setlength{\tabcolsep}{4pt}

\begin{tabular}{llrrrrrrr}
\toprule
Model & Dataset & prefix & $F_{.25}$ & $F_{.5}$ & $F_{.75}$ & $F_{\mathrm{suf}}$ & acc$_{.75}$ & acc$_{\mathrm{suf}}$ \\
\midrule
\multicolumn{9}{l}{\textit{Qwen}}\\
2B & A-OKVQA & 0.820 & 0.063 & 0.183 & 0.293 & 0.263 & 0.663 & 0.680 \\
2B & ScienceQA & 0.833 & 0.063 & 0.123 & 0.197 & 0.223 & 0.760 & 0.750 \\
2B & CV-Bench & 0.847 & 0.153 & 0.250 & 0.300 & 0.367 & 0.673 & 0.590 \\
2B & EuroSAT & 0.500 & 0.587 & 0.403 & 0.430 & 0.390 & 0.490 & 0.540 \\
2B & Pets & 0.720 & 0.310 & 0.333 & 0.500 & 0.607 & 0.430 & 0.383 \\
2B & MMMU & 0.463 & 0.213 & 0.287 & 0.423 & 0.323 & 0.380 & 0.440 \\
4B & A-OKVQA & 0.860 & 0.060 & 0.110 & 0.153 & 0.130 & 0.827 & 0.810 \\
4B & ScienceQA & 0.920 & 0.063 & 0.100 & 0.183 & 0.197 & 0.827 & 0.790 \\
4B & CV-Bench & 0.867 & 0.100 & 0.150 & 0.217 & 0.300 & 0.743 & 0.697 \\
4B & EuroSAT & 0.547 & 0.200 & 0.213 & 0.297 & 0.250 & 0.453 & 0.543 \\
4B & Pets & 0.893 & 0.113 & 0.220 & 0.387 & 0.300 & 0.570 & 0.663 \\
4B & MMMU & 0.580 & 0.137 & 0.220 & 0.333 & 0.247 & 0.497 & 0.547 \\
8B & A-OKVQA & 0.880 & 0.057 & 0.100 & 0.137 & 0.143 & 0.840 & 0.833 \\
8B & ScienceQA & 0.943 & 0.040 & 0.093 & 0.133 & 0.163 & 0.847 & 0.827 \\
8B & CV-Bench & 0.850 & 0.100 & 0.143 & 0.143 & 0.217 & 0.800 & 0.757 \\
8B & EuroSAT & 0.517 & 0.313 & 0.293 & 0.427 & 0.300 & 0.403 & 0.557 \\
8B & Pets & 0.833 & 0.150 & 0.167 & 0.220 & 0.197 & 0.800 & 0.833 \\
8B & MMMU & 0.597 & 0.137 & 0.200 & 0.247 & 0.267 & 0.547 & 0.603 \\
7B & A-OKVQA & 0.900 & 0.060 & 0.123 & 0.170 & 0.133 & 0.820 & 0.847 \\
7B & ScienceQA & 0.853 & 0.090 & 0.103 & 0.127 & 0.170 & 0.817 & 0.803 \\
7B & CV-Bench & 0.820 & 0.130 & 0.230 & 0.270 & 0.320 & 0.713 & 0.673 \\
7B & EuroSAT & 0.353 & 0.653 & 0.550 & 0.777 & 0.437 & 0.143 & 0.300 \\
7B & Pets & 0.927 & 0.113 & 0.187 & 0.457 & 0.390 & 0.513 & 0.570 \\
7B & MMMU & 0.567 & 0.197 & 0.217 & 0.283 & 0.280 & 0.477 & 0.523 \\
\midrule
\multicolumn{9}{l}{\textit{Non-Qwen}}\\
Idefics3-8B & A-OKVQA & 0.860 & 0.160 & 0.380 & 0.677 & 0.717 & 0.307 & 0.227 \\
Idefics3-8B & ScienceQA & 0.880 & 0.087 & 0.317 & 0.423 & 0.430 & 0.583 & 0.527 \\
Idefics3-8B & CV-Bench & 0.717 & 0.263 & 0.497 & 0.553 & 0.507 & 0.453 & 0.440 \\
Idefics3-8B & EuroSAT & 0.347 & 0.483 & 0.753 & 0.793 & 0.787 & 0.140 & 0.127 \\
Idefics3-8B & Pets & 0.590 & 0.337 & 0.457 & 0.537 & 0.997 & 0.200 & 0.023 \\
Idefics3-8B & MMMU & 0.473 & 0.310 & 0.547 & 0.723 & 0.660 & 0.257 & 0.263 \\
LLaVA-OneVision & A-OKVQA & 0.910 & 0.007 & 0.000 & 0.377 & 0.407 & 0.613 & 0.573 \\
LLaVA-OneVision & ScienceQA & 0.967 & 0.007 & 0.027 & 0.227 & 0.243 & 0.770 & 0.743 \\
LLaVA-OneVision & CV-Bench & 0.800 & 0.030 & 0.123 & 0.460 & 0.400 & 0.483 & 0.567 \\
LLaVA-OneVision & EuroSAT & 0.477 & 0.033 & 0.487 & 0.547 & 0.310 & 0.367 & 0.483 \\
LLaVA-OneVision & Pets & 0.707 & 0.067 & 0.343 & 0.593 & 0.907 & 0.307 & 0.080 \\
LLaVA-OneVision & MMMU & 0.577 & 0.053 & 0.277 & 0.467 & 0.363 & 0.367 & 0.440 \\
LLaVA-NeXT & A-OKVQA & 0.843 & 0.037 & 0.013 & 0.320 & 0.657 & 0.617 & 0.337 \\
LLaVA-NeXT & ScienceQA & 0.693 & 0.043 & 0.043 & 0.283 & 0.420 & 0.610 & 0.593 \\
LLaVA-NeXT & CV-Bench & 0.687 & 0.050 & 0.127 & 0.503 & 0.600 & 0.450 & 0.463 \\
LLaVA-NeXT & EuroSAT & 0.497 & 0.167 & 0.167 & 0.610 & 0.717 & 0.203 & 0.120 \\
LLaVA-NeXT & Pets & 0.427 & 0.197 & 0.550 & 0.817 & 0.967 & 0.250 & 0.027 \\
LLaVA-NeXT & MMMU & 0.387 & 0.107 & 0.207 & 0.563 & 0.517 & 0.327 & 0.320 \\
\bottomrule
\end{tabular}
\caption{Token-level sweep for all 42 model--dataset cells. $r$ is the fraction
of the text token count after which the image block is inserted; $r=0.25,0.5,0.75$
are token fractions, not semantic slots, and $r=1$ is suffix. Columns are
prefix accuracy, flips at $r=0.25,0.5,0.75$ and suffix relative to prefix, and
accuracy at $r=0.75$ and suffix.}
\label{tab:token}
\end{table*}
\begin{table*}[t]
\centering
\footnotesize
\resizebox{\textwidth}{!}{
\begin{tabular}{llrrrrrrrrr}
\toprule
Model & Dataset & All-mean & All-logmean & P+S mean & DOF 0.5 & DOF 0.75 & DOF 0.75 flip & $P_{\mathrm{pt}}$ pre & $P_{\mathrm{pt}}$ acc & $P_{\mathrm{pt}}$ flip \\
\midrule
\multicolumn{11}{l}{\textit{Qwen}} \\
2B & A-OKVQA & 0.810 & 0.810 & 0.793 & 0.793 & 0.817 & 0.043 & 0.830 & 0.717 & 0.227 \\
2B & ScienceQA & 0.863 & 0.843 & 0.850 & 0.857 & 0.857 & 0.033 & -- & -- & -- \\
2B & CV-Bench & 0.830 & 0.823 & 0.800 & 0.790 & 0.837 & 0.043 & -- & -- & -- \\
2B & EuroSAT & 0.540 & 0.547 & 0.557 & 0.547 & 0.533 & 0.127 & 0.510 & 0.550 & 0.450 \\
2B & Pets & 0.713 & 0.700 & 0.657 & 0.667 & 0.713 & 0.047 & -- & -- & -- \\
2B & MMMU & 0.473 & 0.477 & 0.477 & 0.483 & 0.470 & 0.087 & -- & -- & -- \\
4B & A-OKVQA & 0.850 & 0.853 & 0.853 & 0.850 & 0.860 & 0.027 & 0.843 & 0.820 & 0.140 \\
4B & ScienceQA & 0.897 & 0.917 & 0.883 & 0.877 & 0.903 & 0.037 & -- & -- & -- \\
4B & CV-Bench & 0.850 & 0.857 & 0.853 & 0.857 & 0.870 & 0.027 & -- & -- & -- \\
4B & EuroSAT & 0.547 & 0.547 & 0.543 & 0.543 & 0.553 & 0.077 & 0.550 & 0.563 & 0.340 \\
4B & Pets & 0.887 & 0.890 & 0.867 & 0.853 & 0.890 & 0.033 & -- & -- & -- \\
4B & MMMU & 0.587 & 0.587 & 0.577 & 0.570 & 0.583 & 0.050 & -- & -- & -- \\
8B & A-OKVQA & 0.880 & 0.877 & 0.863 & 0.867 & 0.873 & 0.037 & 0.867 & 0.830 & 0.133 \\
8B & ScienceQA & 0.937 & 0.937 & 0.927 & 0.933 & 0.943 & 0.013 & -- & -- & -- \\
8B & CV-Bench & 0.853 & 0.857 & 0.853 & 0.853 & 0.857 & 0.033 & -- & -- & -- \\
8B & EuroSAT & 0.600 & 0.593 & 0.567 & 0.580 & 0.580 & 0.123 & 0.537 & 0.507 & 0.407 \\
8B & Pets & 0.870 & 0.897 & 0.870 & 0.877 & 0.880 & 0.060 & -- & -- & -- \\
8B & MMMU & 0.610 & 0.610 & 0.610 & 0.613 & 0.600 & 0.077 & -- & -- & -- \\
7B & A-OKVQA & 0.903 & 0.903 & 0.903 & 0.900 & 0.903 & 0.017 & 0.847 & 0.787 & 0.187 \\
7B & ScienceQA & 0.863 & 0.863 & 0.857 & 0.853 & 0.850 & 0.053 & -- & -- & -- \\
7B & CV-Bench & 0.813 & 0.807 & 0.833 & 0.833 & 0.823 & 0.040 & -- & -- & -- \\
7B & EuroSAT & 0.357 & 0.353 & 0.367 & 0.367 & 0.363 & 0.077 & 0.350 & 0.363 & 0.497 \\
7B & Pets & 0.923 & 0.923 & 0.913 & 0.900 & 0.927 & 0.010 & -- & -- & -- \\
7B & MMMU & 0.570 & 0.563 & 0.567 & 0.563 & 0.560 & 0.053 & -- & -- & -- \\
\midrule
\multicolumn{11}{l}{\textit{Non-Qwen}} \\
Idefics3-8B & A-OKVQA & 0.823 & 0.793 & 0.553 & 0.520 & 0.813 & 0.080 & 0.860 & 0.350 & 0.663 \\
Idefics3-8B & ScienceQA & 0.880 & 0.877 & 0.860 & 0.857 & 0.867 & 0.017 & -- & -- & -- \\
Idefics3-8B & CV-Bench & 0.673 & 0.627 & 0.553 & 0.550 & 0.683 & 0.097 & -- & -- & -- \\
Idefics3-8B & EuroSAT & 0.340 & 0.350 & 0.327 & 0.337 & 0.340 & 0.070 & 0.327 & 0.173 & 0.773 \\
Idefics3-8B & Pets & 0.590 & 0.590 & 0.450 & 0.497 & 0.583 & 0.063 & -- & -- & -- \\
Idefics3-8B & MMMU & 0.477 & 0.460 & 0.397 & 0.387 & 0.477 & 0.107 & -- & -- & -- \\
LLaVA-OneVision & A-OKVQA & 0.917 & 0.917 & 0.913 & 0.910 & 0.910 & 0.000 & 0.883 & 0.677 & 0.320 \\
LLaVA-OneVision & ScienceQA & 0.963 & 0.967 & 0.960 & 0.960 & 0.963 & 0.003 & -- & -- & -- \\
LLaVA-OneVision & CV-Bench & 0.783 & 0.777 & 0.790 & 0.790 & 0.800 & 0.013 & -- & -- & -- \\
LLaVA-OneVision & EuroSAT & 0.483 & 0.473 & 0.477 & 0.477 & 0.473 & 0.047 & 0.523 & 0.450 & 0.270 \\
LLaVA-OneVision & Pets & 0.693 & 0.690 & 0.703 & 0.630 & 0.687 & 0.067 & -- & -- & -- \\
LLaVA-OneVision & MMMU & 0.570 & 0.570 & 0.560 & 0.560 & 0.577 & 0.027 & -- & -- & -- \\
LLaVA-NeXT & A-OKVQA & 0.843 & 0.840 & 0.833 & 0.830 & 0.833 & 0.010 & 0.803 & 0.420 & 0.573 \\
LLaVA-NeXT & ScienceQA & 0.710 & 0.710 & 0.713 & 0.717 & 0.697 & 0.053 & -- & -- & -- \\
LLaVA-NeXT & CV-Bench & 0.690 & 0.690 & 0.677 & 0.680 & 0.683 & 0.097 & -- & -- & -- \\
LLaVA-NeXT & EuroSAT & 0.460 & 0.417 & 0.293 & 0.230 & 0.400 & 0.190 & 0.543 & 0.127 & 0.937 \\
LLaVA-NeXT & Pets & 0.433 & 0.430 & 0.427 & 0.397 & 0.437 & 0.090 & -- & -- & -- \\
LLaVA-NeXT & MMMU & 0.403 & 0.393 & 0.423 & 0.413 & 0.400 & 0.163 & -- & -- & -- \\
\bottomrule
\end{tabular}
}
\caption{Per-cell readout variants behind Table~\ref{tab:fuse}, for all seven
checkpoints, grouped into Qwen and non-Qwen; each row is one checkpoint on one dataset. Acc is the argmax
accuracy of the readout and Flip is its flip relative to prefix, except for the
DOF 0.75 flip column. All-mean is the mean of the four placement
distributions; All-logmean is their log-mean; P+S mean is the prefix--suffix
probability mean. P, Q and O abbreviate prefix, pre-question and pre-options;
P+Q+O, Q+O, P+Q and P+O are means of the named placements, and M+M is the mean
of the all-position mean and all-position log-mean distributions. DOF 0.5 and
DOF 0.75 are dual-order fusion at $\beta=0.5$ and $0.75$. The pointwise columns
are prefix accuracy, suffix accuracy and suffix flip relative to pointwise
prefix; pointwise was run on A-OKVQA and EuroSAT only, and ``--'' marks cells
where it was not run. All values are accuracies except the flip columns. The non-pointwise columns are fused or
averaged readouts, not the raw per-placement accuracies of
Table~\ref{tab:positionsfull}; a fused accuracy can coincide with a placement
accuracy by chance, but the two are different quantities.}
\label{tab:fusefull}
\end{table*}
\begin{table*}[t]
\centering
\scriptsize
\resizebox{\textwidth}{!}{
\begin{tabular}{llrrrrr rrrrr}
\toprule
 & & \multicolumn{5}{c}{Accuracy} & \multicolumn{5}{c}{Flip vs prefix} \\
\cmidrule(lr){3-7}\cmidrule(l){8-12}
Model & Dataset & P+Q+O & Q+O & P+Q & P+O & M+M & P+Q+O & Q+O & P+Q & P+O & M+M \\
\midrule
\multicolumn{12}{l}{\textit{Qwen}} \\
2B & A-OKVQA & 0.823 & 0.817 & 0.823 & 0.820 & 0.810 & 0.030 & 0.053 & 0.010 & 0.050 & 0.067 \\
2B & ScienceQA & 0.863 & 0.860 & 0.857 & 0.870 & 0.857 & 0.050 & 0.093 & 0.047 & 0.070 & 0.057 \\
2B & CV-Bench & 0.850 & 0.833 & 0.847 & 0.840 & 0.830 & 0.040 & 0.090 & 0.013 & 0.077 & 0.090 \\
2B & EuroSAT & 0.530 & 0.537 & 0.500 & 0.547 & 0.547 & 0.130 & 0.190 & 0.073 & 0.177 & 0.193 \\
2B & Pets & 0.713 & 0.707 & 0.717 & 0.710 & 0.707 & 0.040 & 0.093 & 0.020 & 0.077 & 0.057 \\
2B & MMMU & 0.473 & 0.483 & 0.467 & 0.483 & 0.477 & 0.073 & 0.130 & 0.023 & 0.123 & 0.103 \\
4B & A-OKVQA & 0.847 & 0.850 & 0.857 & 0.853 & 0.850 & 0.020 & 0.037 & 0.017 & 0.040 & 0.047 \\
4B & ScienceQA & 0.890 & 0.883 & 0.907 & 0.903 & 0.913 & 0.040 & 0.070 & 0.017 & 0.043 & 0.040 \\
4B & CV-Bench & 0.863 & 0.847 & 0.863 & 0.853 & 0.857 & 0.017 & 0.063 & 0.007 & 0.070 & 0.063 \\
4B & EuroSAT & 0.540 & 0.547 & 0.543 & 0.547 & 0.543 & 0.047 & 0.127 & 0.017 & 0.137 & 0.110 \\
4B & Pets & 0.897 & 0.890 & 0.893 & 0.897 & 0.890 & 0.003 & 0.023 & 0.000 & 0.030 & 0.033 \\
4B & MMMU & 0.593 & 0.603 & 0.580 & 0.590 & 0.593 & 0.037 & 0.077 & 0.023 & 0.073 & 0.060 \\
8B & A-OKVQA & 0.880 & 0.883 & 0.877 & 0.880 & 0.877 & 0.017 & 0.027 & 0.013 & 0.037 & 0.037 \\
8B & ScienceQA & 0.940 & 0.933 & 0.933 & 0.940 & 0.937 & 0.020 & 0.027 & 0.020 & 0.020 & 0.023 \\
8B & CV-Bench & 0.857 & 0.850 & 0.853 & 0.853 & 0.857 & 0.020 & 0.040 & 0.003 & 0.043 & 0.043 \\
8B & EuroSAT & 0.567 & 0.590 & 0.553 & 0.577 & 0.590 & 0.080 & 0.163 & 0.073 & 0.160 & 0.150 \\
8B & Pets & 0.850 & 0.863 & 0.837 & 0.860 & 0.893 & 0.030 & 0.060 & 0.010 & 0.057 & 0.073 \\
8B & MMMU & 0.597 & 0.607 & 0.597 & 0.610 & 0.603 & 0.033 & 0.063 & 0.023 & 0.060 & 0.090 \\
7B & A-OKVQA & 0.910 & 0.897 & 0.903 & 0.907 & 0.903 & 0.033 & 0.047 & 0.020 & 0.033 & 0.037 \\
7B & ScienceQA & 0.860 & 0.853 & 0.873 & 0.860 & 0.863 & 0.053 & 0.077 & 0.053 & 0.057 & 0.073 \\
7B & CV-Bench & 0.823 & 0.800 & 0.823 & 0.803 & 0.813 & 0.053 & 0.083 & 0.010 & 0.080 & 0.077 \\
7B & EuroSAT & 0.370 & 0.370 & 0.370 & 0.370 & 0.357 & 0.097 & 0.183 & 0.053 & 0.147 & 0.170 \\
7B & Pets & 0.927 & 0.930 & 0.927 & 0.927 & 0.927 & 0.007 & 0.010 & 0.007 & 0.007 & 0.013 \\
7B & MMMU & 0.573 & 0.570 & 0.570 & 0.567 & 0.560 & 0.063 & 0.100 & 0.020 & 0.080 & 0.090 \\
\midrule
\multicolumn{12}{l}{\textit{Non-Qwen}} \\
Idefics3-8B & A-OKVQA & 0.850 & 0.823 & 0.863 & 0.823 & 0.807 & 0.040 & 0.093 & 0.010 & 0.087 & 0.080 \\
Idefics3-8B & ScienceQA & 0.883 & 0.857 & 0.877 & 0.877 & 0.880 & 0.037 & 0.073 & 0.027 & 0.050 & 0.043 \\
Idefics3-8B & CV-Bench & 0.703 & 0.653 & 0.720 & 0.660 & 0.650 & 0.050 & 0.160 & 0.010 & 0.157 & 0.167 \\
Idefics3-8B & EuroSAT & 0.347 & 0.367 & 0.353 & 0.360 & 0.343 & 0.110 & 0.207 & 0.020 & 0.200 & 0.190 \\
Idefics3-8B & Pets & 0.583 & 0.597 & 0.573 & 0.590 & 0.587 & 0.027 & 0.043 & 0.030 & 0.047 & 0.040 \\
Idefics3-8B & MMMU & 0.473 & 0.453 & 0.483 & 0.460 & 0.470 & 0.090 & 0.227 & 0.043 & 0.217 & 0.153 \\
LLaVA-OneVision & A-OKVQA & 0.917 & 0.913 & 0.910 & 0.910 & 0.920 & 0.017 & 0.027 & 0.000 & 0.030 & 0.013 \\
LLaVA-OneVision & ScienceQA & 0.963 & 0.963 & 0.963 & 0.967 & 0.963 & 0.020 & 0.020 & 0.013 & 0.017 & 0.020 \\
LLaVA-OneVision & CV-Bench & 0.770 & 0.763 & 0.803 & 0.767 & 0.783 & 0.040 & 0.077 & 0.010 & 0.080 & 0.057 \\
LLaVA-OneVision & EuroSAT & 0.483 & 0.480 & 0.473 & 0.483 & 0.477 & 0.123 & 0.160 & 0.037 & 0.163 & 0.107 \\
LLaVA-OneVision & Pets & 0.687 & 0.690 & 0.700 & 0.690 & 0.693 & 0.023 & 0.033 & 0.013 & 0.027 & 0.013 \\
LLaVA-OneVision & MMMU & 0.563 & 0.560 & 0.583 & 0.557 & 0.570 & 0.083 & 0.143 & 0.017 & 0.133 & 0.083 \\
LLaVA-NeXT & A-OKVQA & 0.847 & 0.843 & 0.843 & 0.847 & 0.843 & 0.023 & 0.037 & 0.013 & 0.027 & 0.020 \\
LLaVA-NeXT & ScienceQA & 0.700 & 0.707 & 0.707 & 0.697 & 0.713 & 0.067 & 0.100 & 0.037 & 0.067 & 0.097 \\
LLaVA-NeXT & CV-Bench & 0.683 & 0.680 & 0.683 & 0.670 & 0.690 & 0.077 & 0.150 & 0.030 & 0.140 & 0.157 \\
LLaVA-NeXT & EuroSAT & 0.507 & 0.497 & 0.507 & 0.500 & 0.443 & 0.023 & 0.033 & 0.010 & 0.017 & 0.130 \\
LLaVA-NeXT & Pets & 0.440 & 0.440 & 0.427 & 0.440 & 0.433 & 0.067 & 0.103 & 0.027 & 0.077 & 0.073 \\
LLaVA-NeXT & MMMU & 0.410 & 0.393 & 0.403 & 0.387 & 0.400 & 0.123 & 0.197 & 0.057 & 0.180 & 0.193 \\
\bottomrule
\end{tabular}
}
\caption{Per-cell averaging ablations behind the averaging rows of
Table~\ref{tab:fuse}, for all seven checkpoints. P is prefix, Q is
pre-question and O is pre-options. P+Q+O averages the three non-suffix
placements, Q+O the two text-side placements, P+Q prefix and pre-question, and
P+O prefix and pre-options; M+M is the mean of the all-position mean and
all-position log-mean distributions. The first five numeric columns are
accuracies; the last five are flip rates relative to prefix.}
\label{tab:avgfull}
\end{table*}
\begin{table*}[t]
\centering
\footnotesize
\resizebox{\textwidth}{!}{
\begin{tabular}{lrrrrrrr}
\toprule
Readout & 2B & 4B & 8B & Qwen2.5-7B & Idefics3 & LLaVA-OV & LLaVA-NeXT \\
\midrule
Prefix & 0.697 / 0.0 & 0.778 / 0.0 & 0.770 / 0.0 & 0.737 / 0.0 & 0.644 / 0.0 & 0.739 / 0.0 & 0.589 / 0.0 \\
Pre-question & 0.702 / 5.7 & 0.769 / 3.2 & 0.776 / 3.3 & 0.742 / 5.1 & 0.641 / 4.1 & 0.736 / 2.3 & 0.595 / 5.4 \\
Pre-options & 0.677 / 18.7 & 0.759 / 12.0 & 0.776 / 11.1 & 0.723 / 15.2 & 0.576 / 29.8 & 0.657 / 20.8 & 0.549 / 21.8 \\
Suffix & 0.564 / 36.2 & 0.675 / 23.7 & 0.735 / 21.4 & 0.619 / 28.8 & 0.268 / 68.3 & 0.481 / 43.8 & 0.310 / 64.6 \\
All-position mean & 0.705 / 8.3 & 0.769 / 5.1 & 0.792 / 5.8 & 0.738 / 7.6 & 0.631 / 8.7 & 0.735 / 4.8 & 0.590 / 11.1 \\
All-position log-mean & 0.700 / 10.5 & 0.775 / 6.2 & 0.795 / 7.2 & 0.736 / 8.1 & 0.616 / 13.4 & 0.732 / 5.6 & 0.580 / 11.6 \\
Prefix+suffix mean & 0.689 / 13.7 & 0.763 / 8.7 & 0.782 / 11.2 & 0.740 / 8.8 & 0.523 / 24.6 & 0.734 / 3.5 & 0.561 / 20.2 \\
P+Q+O mean & 0.709 / 6.1 & 0.772 / 2.7 & 0.782 / 3.3 & 0.744 / 5.1 & 0.640 / 5.9 & 0.731 / 5.1 & 0.598 / 6.3 \\
Q+O mean & 0.706 / 10.8 & 0.770 / 6.6 & 0.788 / 6.3 & 0.737 / 8.3 & 0.625 / 13.4 & 0.728 / 7.7 & 0.593 / 10.3 \\
P+Q mean & 0.702 / 3.1 & 0.774 / 1.3 & 0.775 / 2.4 & 0.744 / 2.7 & 0.645 / 2.3 & 0.739 / 1.5 & 0.595 / 2.9 \\
P+O mean & 0.712 / 9.6 & 0.774 / 6.6 & 0.787 / 6.3 & 0.739 / 6.7 & 0.628 / 12.6 & 0.729 / 7.5 & 0.590 / 8.4 \\
M+M mean & 0.704 / 9.4 & 0.774 / 5.9 & 0.793 / 6.9 & 0.737 / 7.7 & 0.623 / 11.2 & 0.734 / 4.9 & 0.587 / 11.2 \\
Dual-order, $\beta{=}0.5$ & 0.689 / 14.8 & 0.758 / 9.2 & 0.787 / 11.1 & 0.736 / 9.8 & 0.524 / 25.2 & 0.721 / 6.9 & 0.544 / 24.2 \\
Dual-order, $\beta{=}0.75$ & 0.704 / 6.3 & 0.777 / 4.2 & 0.789 / 5.7 & 0.738 / 4.2 & 0.627 / 7.2 & 0.735 / 2.6 & 0.575 / 10.1 \\
\bottomrule
\end{tabular}
}
\caption{Readout results averaged over the six 300-item datasets, one column
per checkpoint. Each cell is accuracy / flip (\%) relative to prefix; P, Q and
O abbreviate prefix, pre-question and pre-options. The per-model means weight
datasets equally, not items. Pointwise is omitted because it covers only
A-OKVQA and EuroSAT; it is reported in Table~\ref{tab:fusefull}. ECE is omitted for
width and is also in Table~\ref{tab:fusefull}.}
\label{tab:bymodel}
\end{table*}
\begin{table*}[t]
\centering
\footnotesize
\resizebox{\textwidth}{!}{
\begin{tabular}{lrrrrrr}
\toprule
Readout & A-OKVQA & ScienceQA & CV-Bench & EuroSAT & Pets & MMMU \\
\midrule
\multicolumn{7}{l}{\textit{All-model mean over seven checkpoints}} \\
Prefix & 0.868 / 0.0 & 0.870 / 0.0 & 0.798 / 0.0 & 0.462 / 0.0 & 0.728 / 0.0 & 0.520 / 0.0 \\
Pre-question & 0.865 / 1.9 & 0.867 / 5.3 & 0.800 / 2.6 & 0.470 / 7.0 & 0.724 / 2.7 & 0.527 / 5.4 \\
Pre-options & 0.826 / 11.0 & 0.819 / 12.7 & 0.680 / 28.1 & 0.484 / 24.9 & 0.730 / 8.9 & 0.503 / 25.2 \\
Suffix & 0.615 / 35.0 & 0.719 / 26.4 & 0.598 / 38.7 & 0.381 / 45.6 & 0.369 / 62.3 & 0.448 / 38.0 \\
All-position mean & 0.861 / 3.7 & 0.873 / 4.7 & 0.785 / 8.0 & 0.475 / 14.2 & 0.730 / 3.4 & 0.527 / 10.0 \\
All-position log-mean & 0.856 / 5.0 & 0.873 / 5.3 & 0.777 / 10.2 & 0.469 / 15.9 & 0.731 / 5.2 & 0.523 / 12.0 \\
Prefix+suffix mean & 0.816 / 9.7 & 0.864 / 6.1 & 0.766 / 13.1 & 0.447 / 20.0 & 0.698 / 12.1 & 0.516 / 16.7 \\
P+Q+O mean & 0.868 / 2.6 & 0.871 / 4.1 & 0.793 / 4.2 & 0.478 / 8.7 & 0.728 / 2.8 & 0.526 / 7.2 \\
Q+O mean & 0.861 / 4.6 & 0.865 / 6.6 & 0.775 / 9.5 & 0.484 / 15.2 & 0.731 / 5.2 & 0.524 / 13.4 \\
P+Q mean & 0.868 / 1.2 & 0.874 / 3.0 & 0.799 / 1.2 & 0.471 / 4.0 & 0.725 / 1.5 & 0.526 / 3.0 \\
P+O mean & 0.863 / 4.3 & 0.873 / 4.6 & 0.778 / 9.2 & 0.483 / 14.3 & 0.730 / 4.6 & 0.522 / 12.4 \\
M+M mean & 0.859 / 4.3 & 0.875 / 5.0 & 0.783 / 9.3 & 0.471 / 15.0 & 0.733 / 4.3 & 0.525 / 11.0 \\
Dual-order, $\beta{=}0.5$ & 0.810 / 10.7 & 0.865 / 6.6 & 0.765 / 13.5 & 0.440 / 23.1 & 0.689 / 15.6 & 0.513 / 17.3 \\
Dual-order, $\beta{=}0.75$ & 0.859 / 3.0 & 0.869 / 3.0 & 0.793 / 5.0 & 0.463 / 10.1 & 0.731 / 5.3 & 0.524 / 8.0 \\
\midrule
\multicolumn{7}{l}{\textit{Qwen mean over four checkpoints}} \\
Prefix & 0.865 / 0.0 & 0.888 / 0.0 & 0.846 / 0.0 & 0.479 / 0.0 & 0.843 / 0.0 & 0.552 / 0.0 \\
Pre-question & 0.863 / 2.2 & 0.882 / 5.7 & 0.846 / 2.0 & 0.492 / 9.3 & 0.846 / 1.8 & 0.556 / 4.8 \\
Pre-options & 0.850 / 7.9 & 0.862 / 9.2 & 0.772 / 19.3 & 0.507 / 25.6 & 0.843 / 7.3 & 0.568 / 16.1 \\
Suffix & 0.792 / 16.8 & 0.792 / 18.8 & 0.679 / 30.1 & 0.485 / 34.4 & 0.613 / 37.3 & 0.528 / 27.9 \\
All-position mean & 0.861 / 4.2 & 0.890 / 4.5 & 0.837 / 6.3 & 0.511 / 14.4 & 0.848 / 3.0 & 0.560 / 7.8 \\
All-position log-mean & 0.861 / 4.9 & 0.890 / 5.3 & 0.836 / 7.1 & 0.510 / 16.2 & 0.853 / 5.2 & 0.559 / 9.2 \\
Prefix+suffix mean & 0.853 / 6.9 & 0.879 / 6.6 & 0.835 / 8.1 & 0.508 / 18.1 & 0.827 / 9.7 & 0.557 / 14.2 \\
P+Q+O mean & 0.865 / 2.5 & 0.888 / 4.1 & 0.848 / 3.2 & 0.502 / 8.8 & 0.847 / 2.0 & 0.559 / 5.2 \\
Q+O mean & 0.862 / 4.1 & 0.883 / 6.7 & 0.833 / 6.9 & 0.511 / 16.6 & 0.848 / 4.7 & 0.566 / 9.2 \\
P+Q mean & 0.865 / 1.5 & 0.892 / 3.4 & 0.847 / 0.8 & 0.492 / 5.4 & 0.843 / 0.9 & 0.553 / 2.2 \\
P+O mean & 0.865 / 4.0 & 0.893 / 4.8 & 0.838 / 6.8 & 0.510 / 15.5 & 0.848 / 4.2 & 0.562 / 8.4 \\
M+M mean & 0.860 / 4.7 & 0.892 / 4.8 & 0.839 / 6.8 & 0.509 / 15.6 & 0.854 / 4.4 & 0.558 / 8.6 \\
Dual-order, $\beta{=}0.5$ & 0.853 / 7.1 & 0.880 / 6.9 & 0.833 / 8.2 & 0.509 / 19.8 & 0.824 / 10.8 & 0.557 / 14.5 \\
Dual-order, $\beta{=}0.75$ & 0.863 / 3.1 & 0.888 / 3.4 & 0.847 / 3.6 & 0.507 / 10.1 & 0.853 / 3.8 & 0.553 / 6.7 \\
\midrule
\multicolumn{7}{l}{\textit{Non-Qwen mean over three checkpoints}} \\
Prefix & 0.871 / 0.0 & 0.847 / 0.0 & 0.734 / 0.0 & 0.440 / 0.0 & 0.574 / 0.0 & 0.479 / 0.0 \\
Pre-question & 0.868 / 1.3 & 0.847 / 4.8 & 0.738 / 3.4 & 0.441 / 4.0 & 0.562 / 3.9 & 0.488 / 6.1 \\
Pre-options & 0.793 / 15.2 & 0.762 / 17.3 & 0.558 / 39.9 & 0.453 / 24.0 & 0.578 / 11.0 & 0.418 / 37.3 \\
Suffix & 0.379 / 59.3 & 0.621 / 36.4 & 0.490 / 50.2 & 0.243 / 60.4 & 0.043 / 95.7 & 0.341 / 51.3 \\
All-position mean & 0.861 / 3.1 & 0.851 / 5.0 & 0.716 / 10.3 & 0.428 / 13.9 & 0.572 / 3.9 & 0.483 / 13.1 \\
All-position log-mean & 0.850 / 5.0 & 0.851 / 5.4 & 0.698 / 14.3 & 0.413 / 15.3 & 0.570 / 5.2 & 0.474 / 15.8 \\
Prefix+suffix mean & 0.767 / 13.4 & 0.844 / 5.4 & 0.673 / 19.9 & 0.366 / 22.6 & 0.527 / 15.3 & 0.460 / 20.0 \\
P+Q+O mean & 0.871 / 2.7 & 0.849 / 4.1 & 0.719 / 5.6 & 0.446 / 8.6 & 0.570 / 3.9 & 0.482 / 9.9 \\
Q+O mean & 0.860 / 5.2 & 0.842 / 6.4 & 0.699 / 12.9 & 0.448 / 13.3 & 0.576 / 6.0 & 0.469 / 18.9 \\
P+Q mean & 0.872 / 0.8 & 0.849 / 2.6 & 0.736 / 1.7 & 0.444 / 2.2 & 0.567 / 2.3 & 0.490 / 3.9 \\
P+O mean & 0.860 / 4.8 & 0.847 / 4.4 & 0.699 / 12.6 & 0.448 / 12.7 & 0.573 / 5.0 & 0.468 / 17.7 \\
M+M mean & 0.857 / 3.8 & 0.852 / 5.3 & 0.708 / 12.7 & 0.421 / 14.2 & 0.571 / 4.2 & 0.480 / 14.3 \\
Dual-order, $\beta{=}0.5$ & 0.753 / 15.4 & 0.844 / 6.1 & 0.673 / 20.6 & 0.348 / 27.6 & 0.508 / 21.9 & 0.453 / 21.0 \\
Dual-order, $\beta{=}0.75$ & 0.852 / 3.0 & 0.842 / 2.4 & 0.722 / 6.9 & 0.404 / 10.2 & 0.569 / 7.3 & 0.484 / 9.9 \\
\bottomrule
\end{tabular}
}
\caption{Readout results averaged over all seven (top), the four Qwen (middle)
or the three non-Qwen (bottom) checkpoints, one column per dataset. Each cell
is accuracy / flip (\%) relative to prefix; P, Q and O abbreviate prefix,
pre-question and pre-options. Checkpoints are weighted equally, not items, so
the all-model panel gives the Qwen family four of seven parts.
Pointwise is
omitted because it covers only A-OKVQA and EuroSAT, and ECE is omitted for
width; both are in Table~\ref{tab:fusefull}.}
\label{tab:bydataset}
\end{table*}

\begin{table*}[t]
\centering
\scriptsize
\resizebox{0.8\textwidth}{!}{
\begin{tabular}{llrrrr}
\toprule
Model & Dataset & $L_{pre}$ & $L_{suf}$ & $P_{id}$ & $P_{perm}$ \\
\midrule
Qwen3-VL-2B & A-OKVQA & 0.053 & 0.160 & 0.263 & 0.300 \\
Qwen3-VL-2B & CV-Bench & 0.193 & 0.477 & 0.367 & 0.370 \\
Qwen3-VL-2B & EuroSAT & 0.310 & 0.343 & 0.390 & 0.470 \\
Qwen3-VL-2B & MMMU & 0.263 & 0.350 & 0.323 & 0.343 \\
Qwen3-VL-2B & Pets & 0.167 & 0.547 & 0.607 & 0.610 \\
Qwen3-VL-2B & ScienceQA & 0.090 & 0.177 & 0.223 & 0.190 \\
Qwen3-VL-4B & A-OKVQA & 0.043 & 0.137 & 0.130 & 0.150 \\
Qwen3-VL-4B & CV-Bench & 0.047 & 0.237 & 0.300 & 0.407 \\
Qwen3-VL-4B & EuroSAT & 0.150 & 0.143 & 0.250 & 0.290 \\
Qwen3-VL-4B & MMMU & 0.193 & 0.310 & 0.247 & 0.317 \\
Qwen3-VL-4B & Pets & 0.047 & 0.153 & 0.300 & 0.277 \\
Qwen3-VL-4B & ScienceQA & 0.063 & 0.153 & 0.197 & 0.233 \\
Qwen3-VL-8B & A-OKVQA & 0.060 & 0.073 & 0.143 & 0.133 \\
Qwen3-VL-8B & CV-Bench & 0.027 & 0.180 & 0.217 & 0.280 \\
Qwen3-VL-8B & EuroSAT & 0.140 & 0.150 & 0.300 & 0.313 \\
Qwen3-VL-8B & MMMU & 0.187 & 0.193 & 0.267 & 0.267 \\
Qwen3-VL-8B & Pets & 0.127 & 0.107 & 0.197 & 0.173 \\
Qwen3-VL-8B & ScienceQA & 0.043 & 0.110 & 0.163 & 0.160 \\
Qwen2.5-VL-7B & A-OKVQA & 0.033 & 0.067 & 0.133 & 0.160 \\
Qwen2.5-VL-7B & CV-Bench & 0.067 & 0.133 & 0.320 & 0.327 \\
Qwen2.5-VL-7B & EuroSAT & 0.187 & 0.293 & 0.437 & 0.417 \\
Qwen2.5-VL-7B & MMMU & 0.180 & 0.247 & 0.280 & 0.323 \\
Qwen2.5-VL-7B & Pets & 0.050 & 0.280 & 0.390 & 0.307 \\
Qwen2.5-VL-7B & ScienceQA & 0.077 & 0.080 & 0.170 & 0.177 \\
Idefics3-8B & A-OKVQA & 0.107 & 0.797 & 0.717 & 0.690 \\
Idefics3-8B & CV-Bench & 0.270 & 0.923 & 0.507 & 0.437 \\
Idefics3-8B & EuroSAT & 0.457 & 0.923 & 0.787 & 0.813 \\
Idefics3-8B & MMMU & 0.343 & 0.723 & 0.660 & 0.650 \\
Idefics3-8B & Pets & 0.223 & 0.973 & 0.997 & 0.957 \\
Idefics3-8B & ScienceQA & 0.067 & 0.603 & 0.430 & 0.470 \\
LLaVA-OneVision-7B & A-OKVQA & 0.023 & 0.303 & 0.407 & 0.423 \\
LLaVA-OneVision-7B & CV-Bench & 0.080 & 0.433 & 0.400 & 0.450 \\
LLaVA-OneVision-7B & EuroSAT & 0.453 & 0.507 & 0.310 & 0.400 \\
LLaVA-OneVision-7B & MMMU & 0.167 & 0.377 & 0.363 & 0.417 \\
LLaVA-OneVision-7B & Pets & 0.170 & 0.870 & 0.907 & 0.843 \\
LLaVA-OneVision-7B & ScienceQA & 0.007 & 0.200 & 0.243 & 0.227 \\
LLaVA-NeXT-7B & A-OKVQA & 0.050 & 0.570 & 0.657 & 0.593 \\
LLaVA-NeXT-7B & CV-Bench & 0.107 & 0.817 & 0.600 & 0.520 \\
LLaVA-NeXT-7B & EuroSAT & 0.400 & 0.567 & 0.717 & 0.703 \\
LLaVA-NeXT-7B & MMMU & 0.260 & 0.513 & 0.517 & 0.450 \\
LLaVA-NeXT-7B & Pets & 0.410 & 0.957 & 0.967 & 0.877 \\
LLaVA-NeXT-7B & ScienceQA & 0.127 & 0.337 & 0.420 & 0.377 \\
\bottomrule
\end{tabular}
}
\caption{Per-cell label-permutation 2$\times$2: $L$ are identity-vs-permuted
letter flips at prefix/suffix; $P$ are suffix-vs-prefix position flips under
identity/permuted letters. Option order is fixed.}
\label{tab:labelcells}
\end{table*}

\end{document}